%% file: main.tex
\documentclass[runningheads]{llncs}

\usepackage{eccv}

\usepackage{eccvabbrv}

\usepackage{graphicx}
\usepackage{booktabs}
\usepackage{multirow}

\usepackage{algorithm}
\usepackage{algpseudocode}
\usepackage{array}

\usepackage{subcaption}
\usepackage[accsupp]{axessibility}  

\usepackage{hyperref}

\usepackage{orcidlink}

\begin{document}

\title{Score-Based Matching with Target Guidance for Cryo-EM Denoising} 





\author{Xiaoqi Wu\inst{1}\textsuperscript{$\dagger$} \and
Xueying Zhan\inst{1}\textsuperscript{$\dagger$} \and
Wen Li\inst{1} \and
Junhao Wu\inst{2} \and
Xin Huang\inst{2} \and
Min Xu\inst{1}\thanks{Corresponding author.}}
\authorrunning{X.~Wu et al.}
\institute{
Carnegie Mellon University \and
Towson University\\
\email{\{xiaoqiw, xueyingz, wenli3, mxu1\}@andrew.cmu.edu \\
jwu17@students.towson.edu\\
xhuang@towson.edu}
}

\maketitle
\renewcommand{\thefootnote}{\fnsymbol{footnote}}
\footnotetext[4]{Xiaoqi Wu and Xueying Zhan contributed equally to this work.}

\begin{abstract}
Cryo-electron microscopy (cryo-EM) enables single-particle analysis of biological macromolecules under strict low-dose imaging conditions, but the resulting micrographs often exhibit extremely low signal-to-noise ratios and weak particle visibility. Image denoising is therefore an important preprocessing step for downstream cryo-EM analysis, including particle picking, 2D classification, and 3D reconstruction. Existing cryo-EM denoising methods are commonly trained with pixel-wise or Noise2Noise-style objectives, which can improve visual quality but do not explicitly account for structural consistency required by downstream analysis.
In this work, we propose a score-based denoising framework for cryo-EM that learns the clean-data score to recover particle signals while better preserving structural information. Building on this formulation, we further introduce a target-guided variant that incorporates reference-density guidance to stabilize score learning under weak and ambiguous signal conditions. Rather than simply amplifying particle-like responses, our framework better suppresses structured low-frequency background, which improves particle--background separability for downstream analysis.
Experiments on multiple cryo-EM datasets show that our score-based methods consistently improve downstream particle picking and produce more structure-consistent 3D reconstructions. Experiments on multiple cryo-EM datasets show that our methods improve downstream particle picking and produce more structure-consistent reconstructions. 
\keywords{Cryo-electron Microscopy \and Image Denoising \and Score-based Models}
\end{abstract}

\section{Introduction}
\label{sec:intro}

\begin{figure}[t]
  \centering
  \includegraphics[width=0.9\textwidth]{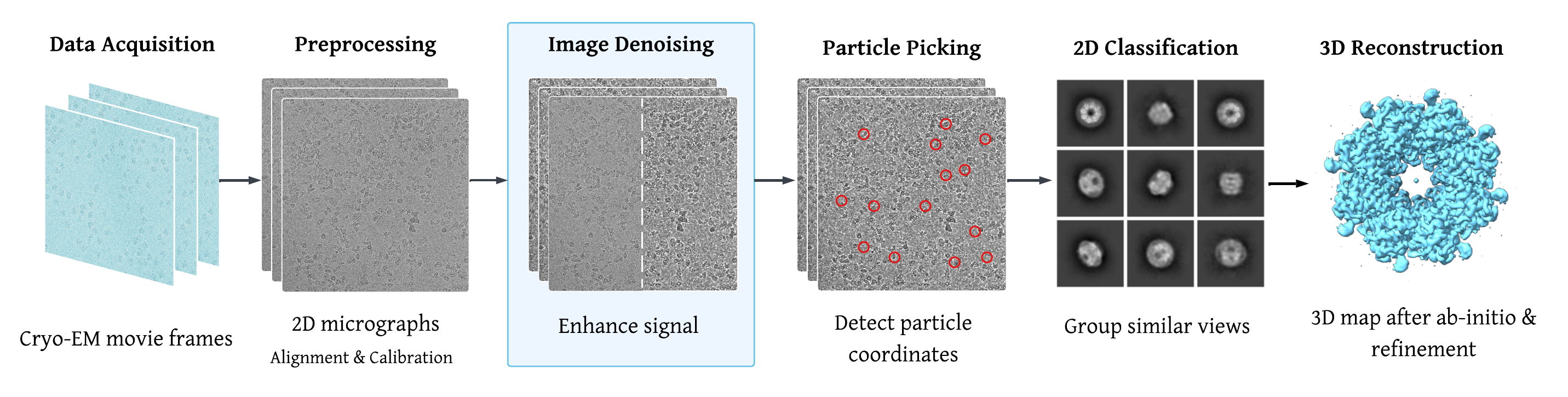}
  \caption{
Cryo-EM single-particle analysis pipeline, illustrating the main computational stages from raw cryo-EM micrographs to 3D reconstructed particle density map.
}
  \label{fig:cryoem-workflow}
\end{figure}

Cryo-EM enables the reconstruction of 3D structures of biological macromolecules from large collections of noisy 2D projection images acquired under cryogenic conditions~\cite{Milne2013CryoEMPrimer,Kuhlbrandt2014ResolutionRevolution,Nogales2016CryoEM}. 
Due to strict low-dose imaging requirements, cryo-EM micrographs exhibit extremely low SNR and weak image contrast. In single-particle analysis (SPA), these micrographs are processed by a multi-stage pipeline including preprocessing, particle picking, 2D classification, and 3D reconstruction, where image-level errors propagate and affect downstream structural recovery~\cite{Zhang2023GENEM}. Fig.~\ref{fig:cryoem-workflow} summarizes this workflow. Improving micrograph quality therefore remains critical for reliable cryo-EM analysis. Image denoising is particularly challenging because particle signals are weak, spatially small, and often indistinguishable from structured background noise, especially for small particles or low-contrast views where aggressive denoising may remove meaningful structural information. Unlike natural image restoration tasks, cryo-EM denoising lacks clean ground-truth supervision due to radiation damage constraints. Moreover, modern restoration methods increasingly rely on large-scale pretrained models and natural-image priors, which transfer poorly to cryo-EM data due to the substantial domain gap. Consequently, general-purpose vision denoisers often struggle to preserve weak particle structures under extreme noise conditions.

Early cryo-EM denoising mainly used classical signal-processing methods such as Gaussian filtering~\cite{Frank2006ThreeDEM}, but the gains are limited in the extreme low-SNR regime. Recent work has therefore shifted toward learning-based denoisers trained directly on cryo-EM data. 
Many existing methods adopt Noise2Noise (N2N)-style training, enabling self-supervised learning from paired noisy observations without clean ground truth~\cite{N2N,topaz,NT2C}. While improving visual quality, these approaches rely on pixel-level regression objectives that do not explicitly enforce structural consistency, often leading to over-smoothing or structural distortion under weak particle signals. Methods like DRACO improve training stability through modified data splitting and masking strategies~\cite{draco}. However, two challenges remain: regression objectives do not explicitly model noise statistics in low-SNR regimes, and improved denoising quality does not necessarily translate into better downstream reconstruction. This reveals a gap between image restoration objectives and structural recovery requirements.

Motivated by these challenges, we adopt denoising score matching (DSM) as the foundation of our approach, since score-based learning models image corruption through the underlying data distribution rather than direct pixel-wise regression, making it more suitable for preserving structural signals under heavy noise without clean supervision. 
However, under extremely low SNR, score estimation becomes ill-conditioned in the low-noise regime close to the clean signal, where weak structural cues provide limited optimization guidance. DSM alone therefore fails to recover reliable score directions. To address this issue, we then introduce target-guided score learning by incorporating prior particle-structure information in the form of a reference density map. The proposed guidance stabilizes score estimation near the clean signal manifold, providing directional cues under severe noise and improving preservation of fine particle details.
Our contributions are summarized as follows:
\begin{itemize}
\item We analyze instability in score-based denoising for cryo-EM under challenging noise conditions, and propose a target-guided score-based framework that preserves particle-relevant structure without requiring clean ground truth micrographs as supervision.
\item We introduce similarity-guided annealing, an adaptive training scheme that balances target guidance and score matching to preserve particle structure without over-suppressing non-particle regions.
\item Experiments across multiple cryo-EM datasets show consistent gains in downstream tasks like particle picking and improved 3D reconstruction quality.
\end{itemize}

\section{Related Work}
\label{sec:related_work}
\paragraph{\textbf{Traditional Cryo-EM Denoising Methods.}}
Early cryo-EM denoising methods relied on classical signal-processing techniques such as Gaussian and Wiener filtering~\cite{jiang2025review}. They assume simplified noise models and handcrafted priors, which provide limited improvements under extremely low-SNR conditions and often fail to recover fine structural details required for downstream structural analysis.

\paragraph{\textbf{Self-supervised Denoising with Noisy Observations.}}
Deep learning methods have significantly advanced cryo-EM denoising. N2N~\cite{N2N} enables self-supervised learning from paired noisy observations without clean ground truth. Topaz-Denoise~\cite{topaz} adopts this paradigm using a U-Net architecture and demonstrates strong robustness across datasets. More recently, foundation-model-style pretraining has been explored for cryo-EM data, such as DRACO~\cite{draco}, which combines self-supervised pretraining with N2N-based learning. Despite their empirical success, these approaches rely on regression-based objectives that implicitly learn signal statistics without explicitly enforcing structural detail preserving, which lead to structural distortion under weak particle signals.

\paragraph{\textbf{Blind-spot and Pseudo-supervised Approaches.}}
Blind-spot approaches such as Noise2Void~\cite{Noise2void} and Self2Self~\cite{self2self} enable denoising from single noisy observations by predicting masked pixels from local context. However, their locality assumption limits the modeling of global structural information, which is critical in cryo-EM images with weak and spatially correlated signals. Pseudo-supervised methods have also been explored; for example, NT2C~\cite{NT2C} learns noise characteristics using simulated clean data. These synthetic supervisions introduce a statistical mismatch between simulated and experimental noise distributions, limiting generalization to real micrographs.

\paragraph{\textbf{Score-based Models for Cryo-EM Denoising.}}
Score-based generative models estimate data distributions through score learning~\cite{scorebased}. DSM~\cite{DSM} provides a principled framework for modeling noise corruption without clean supervision. However, under the extremely low-SNR conditions of cryo-EM imaging, score estimation can become unstable near the clean signal regime, leading to loss of subtle structural details. Target Score Matching~\cite{TargetScoreMatching2024} extends DSM by incorporating task-preferred properties into score estimation. Our work builds on this line by introducing structure-guided score learning tailored to cryo-EM denoising.

\begin{figure}[tb]
  \centering
  \includegraphics[width=\textwidth]{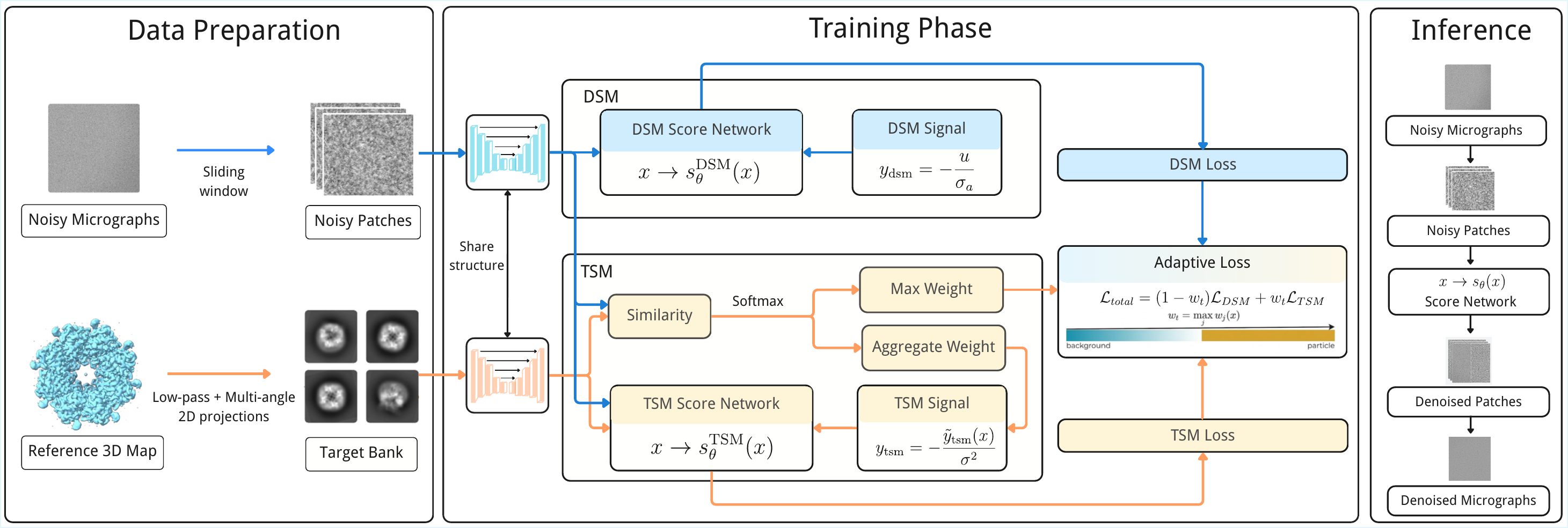}
  \caption{
    Overview of the proposed target-guided score-based denoising framework.
  }
  \label{fig:overview}
\end{figure}

\section{Methodology}
\label{sec:methodology}
We first summarize the cryo-EM noise characteristics and then introduce our target-guided score-based denoising framework.

\subsection{Cryo-EM Noise Pattern}
Cryo-EM micrographs are acquired under dose-limited imaging, where the dominant noise sources are commonly categorized as: (i) \emph{structural noise} (irreproducible background and conformational variability), (ii) \emph{shot noise} (electron counting statistics), and (iii) \emph{detector/digitization noise}~\cite{baxter2009determination,NT2C}.
We summarize these effects using an effective additive formulation:
\begin{equation}
    X = Y + \varepsilon, \qquad 
    \varepsilon=\varepsilon_{\text{struct}}+\varepsilon_{\text{shot}}+\varepsilon_{\text{det}}.
\end{equation}

Shot noise is signal-dependent, modeled as 
$X_i \sim \mathrm{Poisson}(\alpha Y_i + b)$. Detector noise is typically approximated as zero-mean Gaussian. For tractability, we use a single isotropic Gaussian surrogate in both analysis and experiments.

\subsection{Problem Definition}
\label{sec:problem_def}
We study denoising of cryo-EM \emph{micrographs}, i.e., large grayscale microscope images stored in scientific formats (e.g., \texttt{MRC}, \texttt{TIFF}). 
Since micrographs are typically several thousand pixels per side (e.g., $4096\times4096$), we perform learning on local patches. 
Let $x\in\mathbb{R}^{h\times w}$ be a noisy patch and $y\in\mathbb{R}^{h\times w}$ its (unobserved) clean counterpart. 
We vectorize patches and write $x,y\in\mathbb{R}^{d}$ with $d=h\cdot w$. 
In cryo-EM, clean ground truth $y$ is generally unavailable.

Let $X\sim p_X$ and $Y\sim p_Y$ denote the noisy and clean patch-level random variables. 
Define the conditional score of the corruption model as $s_{X\mid Y}(x\mid y):=\nabla_x \log p_{X\mid Y}(x\mid y)$. 
The noisy score satisfies
\begin{equation}
    \nabla_x \log p_{X}(x)=\int p_{Y\mid X}(y\mid x)\, s_{X\mid Y}(x \mid y)\, \mathrm{d} y ,
\end{equation}
where $p_{Y\mid X}(y\mid x)$ denotes the posterior distribution of the latent clean patch given $X=x$. Intuitively, this identity expresses the score of the noisy-patch distribution $p_X$ as a posterior expectation of the noise-model score under $p_{Y\mid X}$.

When the noise model is known and paired samples $(X,Y)$ are available, the score $\nabla_x \log p_X(x)$ can be approximated by a score network $s_X^\theta(x)$ by minimizing
\begin{equation}
	\ell(\theta)
	=
	\mathbb{E}_{X,Y}
	\big\| s_X^\theta(X) - s_{X\mid Y}(X\mid Y) \big\|^2,
\end{equation}
which yields the DSM objective~\cite{scorebased}.
Even under additive Gaussian corruption $X=Y+N$ with $N\sim\mathcal{N}(0,\sigma_a^2 I)$, the score target becomes ill-conditioned as $\sigma_a$ decreases. 
Specifically,
$s_{X\mid Y}(x\mid y)=\nabla_x \log p_{X\mid Y}(x\mid y)=-(x-y)/\sigma_a^2$,
so its scale grows as $\sigma_a^{-2}$. Consequently, the posterior variance satisfies
\begin{equation}
\sum_{i=1}\nolimits^{d} \operatorname{Var}_{Y \mid X}\!\left(\left(s_{X\mid Y}(X \mid Y)\right)_i\right) \sim d\, \sigma_a^{-2}\quad \text{as }\sigma_a \to 0,
\end{equation}
leading to unstable score estimation in the low-noise regime relevant to cryo-EM.
Although denoising should become easier as noise decreases, the DSM target becomes ill-conditioned as $\sigma_a \to 0$. 
This is critical for cryo-EM, where informative structures concentrate near the clean manifold; Fig.~\ref{fig:overview} summarizes our target-guided score-based denoising pipeline.

\subsection{Optimization With Target Guidance}
\label{sec:optim}

Motivated by the instability of denoising score matching in the low-noise regime, we reformulate score estimation by exploiting a key property of additive Gaussian corruption. 
When the corruption noise is sufficiently small, the score of the noisy observation approaches that of the clean signal, i.e.,
$s_X(x)=\nabla_x \log p_X(x)\approx \nabla_x\log p_Y(x)$.
This observation enables incorporating \emph{target-side} structural information to guide score estimation, instead of relying solely on the conditional noise-model score
$s_{X\mid Y}(x\mid y)=\nabla_x\log p_{X\mid Y}(x\mid y)$.
Following target score matching~\cite{TargetScoreMatching2024}, the marginal score admits the approximation
\begin{equation}
\nabla_x \log p_X(x)
\approx
\mathbb{E}\!\left[\nabla_y \log p_Y(Y)\mid X=x\right].
\end{equation}

In practice, we approximate the clean-data score using a target-score network $s_Y^\phi(y)$ and train the noisy score estimator $s_X^\theta$ to match this target-guided estimate. 
Unlike DSM, whose target magnitude scales with $\sigma_a^{-2}$ under Gaussian noise, the proposed formulation avoids explicit noise amplification and provides a substantially more stable training signal in the low-noise regime.

\subsection{Target Matching}
\label{sec:target_matching}

To approximate the target-side expectation in Sec.~\ref{sec:optim}, we construct a bank of coarse structural targets representing object-consistent particle appearances. 
When the particle identity in SPA is known, publicly available 3D templates from EMDB~\cite{wwpdb2024emdb} or PDB~\cite{burley2017protein}\footnote{\texttt{EMDB}: \url{https://www.ebi.ac.uk/emdb/}; \texttt{PDB}: \url{https://www.rcsb.org/}.} can be used as reference volumes. 
When no external structural template is available, a coarse reference can alternatively be estimated directly from the data (e.g., via ab-initio reconstruction), serving the same role as a low-frequency structural prior. 
Since such references often contain high-frequency details not reliably observable in noisy cryo-EM micrographs, we apply low-pass filtering to retain only coarse structural components.

From the reference volume, we generate a bank of 2D target projections $\{\tilde{y}_j\}_{j=1}^{m}$ over multiple viewing directions, where each $\tilde{y}_j\in\mathbb{R}^{d}$ is a plausible particle observation. 
Let $\psi(\cdot):\mathbb{R}^{d}\rightarrow\mathbb{R}^{q}$ be a feature mapping and define $z_j=\psi(\tilde{y}_j)\in\mathbb{R}^{q}$. 
These features span a target subspace
$\mathcal{V}_T=\mathrm{span}\{z_1,\ldots,z_m\}\subset\mathbb{R}^{q}$,
with orthogonal projector $B_T$. 
Given a noisy patch $x\in\mathbb{R}^{d}$ and its feature representation $x'=\psi(x)$, we project it onto the target subspace as $\hat{x}' = B_T x'$. 
This projection preserves components consistent with the target manifold while suppressing orthogonal variations, providing a feature-space approximation of the target-consistent component required in Sec.~\ref{sec:optim}.

\subsection{Train with Target-Score Supervision}
\label{sec:target_score_supervision}
We use a surrogate target score as a supervision signal to train the noisy score network
$s_\theta(x) \approx \nabla_x \log p_X(x)$.
Consistent with the additive Gaussian corruption model used during training, we introduce an isotropic Gaussian \emph{surrogate score model}
$p_Y^{\text{sur}}(y)=\mathcal{N}(0,\sigma^2 I)$ solely to define a closed-form target score field, where $\sigma^2$ is estimated once from the target bank (empirical Bayes), and targets are centered relative to the matched structural reference.
This surrogate does not model the true clean-data distribution and is not used to generate targets; it only provides an analytic score field 
$\nabla_y \log p_Y^{\text{sur}}(y) = -y/\sigma^2$.

We train the noisy score network $s_\theta(x)\approx \nabla_x \log p_X(x)$ using a surrogate target score as supervision.
Under the additive Gaussian corruption model, we introduce an isotropic Gaussian surrogate
$p_Y^{\text{sur}}(y)=\mathcal{N}(0,\sigma^2 I)$
to define a closed-form target score field, where $\sigma^2$ is estimated once from the target bank (empirical Bayes) and targets are centered relative to the matched structural reference.
This surrogate is used only for the analytic score
$\nabla_y \log p_Y^{\text{sur}}(y) = -y/\sigma^2$,
not to model the true clean-data distribution or generate targets.

\subsubsection{Construct TSM pairs}

For $x\in\mathbb{R}^{d}$, we build an input-adaptive target to approximate the conditional expectation in Sec.~\ref{sec:optim}. 
Given target features $\{z_j\}_{j=1}^m$ and the projected noisy feature $\hat{x}'$ (Sec.~\ref{sec:target_matching}), we compute cosine similarities
\[
a_j(x)=\mathrm{sim}(\hat{x}',z_j)
=\frac{\hat{x}'^\top z_j}{\|\hat{x}'\|_2\,\|z_j\|_2},\qquad j=1,\ldots,m.
\]
We convert them into weights via a temperature-scaled softmax,
\begin{equation}
w_j(x)=\frac{\exp(a_j(x)/\tau)}{\sum_{k=1}^{m}\exp(a_k(x)/\tau)},\qquad \sum\nolimits_{j=1}^{m}w_j(x)=1,
\end{equation}
and form the customized target by weighted aggregation:
\begin{equation}
\tilde{y}_{\mathrm{tsm}}(x)=\sum\nolimits_{j=1}^{m} w_j(x)\,\tilde{y}_j.
\end{equation}
This yields the training pair $(x,\tilde{y}_{\mathrm{tsm}}(x))$ and can be viewed as a nonparametric estimate of the conditional target expectation in Sec.~\ref{sec:optim}.

\subsubsection{Rescaled objective under AR-DAE parameterization}

Replacing the conditional noise-model score with target-guided supervision introduces scale mismatch across noise levels. 
To stabilize optimization, we follow the AR-DAE preconditioning analysis in~\cite{lim2020ardae,scorebased,TargetScoreMatching2024} and use the following per-sample rescaled loss:
{
\begin{equation}
\begin{aligned}
\ell_{\text{TSM}}^{\text{res}}(x;\theta)
&:= \tfrac{\sigma^2 \sigma_a^2 + \sigma^4}{\sigma^2}
\Bigg\|
\tfrac{\sigma_a}{\sigma\sqrt{\sigma_a^2+\sigma^2}}
s_{\theta}(\tfrac{x}{\sqrt{\sigma^2+\sigma_a^2}}+\sigma_a)
+ \tfrac{x}{\sigma^2+\sigma_a^2}
- \tfrac{\tilde{y}_{\mathrm{tsm}}(x)}{\sigma^2}
\Bigg\|_2^2 .
\end{aligned}
\end{equation}
}
We optimize the expected objective
$L_{\text{TSM}}^{\text{res}}(\theta)=\mathbb{E}_{x}[\ell_{\text{TSM}}^{\text{res}}(x;\theta)]$.
This rescaling preserves target-score supervision while restoring scale compatibility with DSM, leading to more stable gradients, especially at low noise. 
As in standard score-based training, $\sigma_a$ can be gradually annealed during optimization.

\subsubsection{Relation to DSM}
Under additive Gaussian corruption,
\begin{equation}
x = y + \sigma_a u, \quad u \sim \mathcal{N}(0, I),
\end{equation}
the noise-model score is
$\nabla_x \log p_{X\mid Y}(x \mid y) = -\frac{x - y}{\sigma_a^2}$.
The corresponding per-sample DSM loss is
\begin{equation}
\ell_{\text{DSM}}(x,u;\theta)
= \left\| s_\theta(x+\sigma_a u) + \frac{u}{\sigma_a} \right\|_2^2
= \left\| \sigma_a\, s_\theta(x + \sigma_a u) + u \right\|_2^2,
\end{equation}
where the second form corresponds to the AR-DAE residual parameterization. 
The expected DSM objective is
\[
L_{\text{DSM}}(\theta)=\mathbb{E}_{x,u}\big[\ell_{\text{DSM}}(x,u;\theta)\big].
\]
\emph{DSM and TSM estimate the same marginal score $\nabla_x \log p_X(x)$ but use different supervision signals.} 
TSM can be viewed as a variance-reduced approximation to DSM, replacing the conditional noise-model score with a target-guided estimate under the small-noise assumption.

\subsection{Adaptive Learning}

DSM and TSM provide complementary supervision signals during training. 
DSM captures globally consistent image statistics dominated by background and low-SNR components, while TSM enhances particle-specific structural information through target guidance.
The reliability of target-guided supervision varies across noisy observations. 
While DSM remains stable regardless of particle visibility, TSM depends on alignment between the noisy input and proxy structural targets. 
We therefore introduce an adaptive learning strategy that interpolates between DSM and TSM according to target-alignment confidence.
Using the similarity weights $\{w_j(x)\}_{j=1}^{m}$ defined in Sec.~\ref{sec:target_score_supervision}, we measure alignment confidence as
\begin{equation}
w_t(x) := \max_{1\le j\le m} w_j(x)\in[0,1],
\end{equation}
which reflects consistency between the noisy observation and at least one target projection under unknown viewing directions in SPA.
The confidence weight $w_t(x)$ is computed without gradient propagation and used only as a sample-wise weighting factor. 
The final objective interpolates between TSM and DSM:
\begin{equation}
L_{\text{adp}}(\theta)
= \mathbb{E}_{x}\Big[
w_t(x)\,\ell_{\mathrm{TSM}}^{\text{res}}(x;\theta)
+
(1-w_t(x))\,\ell_{\mathrm{DSM}}(x;\theta)
\Big].
\end{equation}

Samples with confident alignment emphasize target-guided learning, while uncertain inputs (non-particle regions) naturally fall back to DSM supervision, improving robustness under heterogeneous cryo-EM imaging conditions.

\subsection{Training Procedure}
\label{sec:training}

Fig.~\ref{fig:cryoem-workflow} summarizes the overall target-guided training pipeline.
Algorithm~\ref{alg:tsm_train} provides the detailed procedure.
During training, noisy cryo-EM micrographs are partitioned into patches using a sliding-window strategy.
At iteration $t$, a noise level $\sigma_a(t)$ and a global guidance coefficient $\lambda(t)$ are set according to predefined schedules, with $\lambda(t)=0$ during an initial warmup phase to stabilize early training.
For each patch $x$, similarities to the target bank are computed to construct an input-adaptive proxy target $\tilde{y}_{\mathrm{tsm}}(x)$ and a confidence score $w_t(x)$.
The effective supervision weight is given by $\lambda(t) w_t(x)$, yielding the combined loss
\[
\ell_{final} = \lambda(t) w_t(x)\,\ell_{\mathrm{TSM}} 
      + \big(1-\lambda(t) w_t(x)\big)\,\ell_{\mathrm{DSM}}.
\]
When alignment confidence is low or during early iterations, optimization relies primarily on DSM.
As training progresses and alignment improves, target-guided supervision gradually dominates.

\begin{algorithm}[t]
\caption{Target-Guided Score Training}
\label{alg:tsm_train}
\begin{algorithmic}[1]

\Require Noisy micrographs; target bank $\{\tilde{y}_j\}_{j=1}^m$
\Ensure Trained score network $s_\theta$

\State Estimate $\sigma^2$ from target bank
\State Extract noisy patches $\{x\}$
\For{$t=1$ to $T$}
    \State Set $\sigma_a(t)$ and $\lambda(t)$
    \ForAll{$x$}
        \State Compute $\psi_\theta(x)$
        \State Compute similarities $\{a_j(x)\}$
        \State Compute weights $w_j(x)$
        \State $\tilde{y}_{\mathrm{tsm}}(x) \gets \sum_j w_j(x)\tilde{y}_j$
        \State $w_t(x) \gets \max_j w_j(x)$
        \State $\ell_{final} \gets \lambda(t)w_t(x)\ell_{\mathrm{TSM}}
              + (1-\lambda(t)w_t(x))\ell_{\mathrm{DSM}}$
    \EndFor
    \State Update $\theta$ via SGD
\EndFor \\
\Return $s_\theta$

\end{algorithmic}
\end{algorithm}

\section{Experiments}
We evaluate our method on multiple cryo-EM datasets against representative baselines. This section first introduces the experimental setup and then reports results on denoising, particle picking, and 3D reconstruction.
\subsection{Experimental Setup}
\paragraph{Datasets.}
We select cryo-EM datasets from Electron Microscopy Public Image Archive (EMPIAR)\footnote{\texttt{EMPIAR}: \url{https://www.ebi.ac.uk/empiar/}.}~\cite{iudin2023empiar}, including EMPIAR-10081~\cite{lee2017structures}, EMPIAR-10289, and EMPIAR-10291~\cite{burendei2020cryo}. 
For all datasets, we adopt the CryoPPP benchmark~\cite{dhakal2023cryoppp}, which provides standardized preprocessing micrographs (e.g., motion correction) and expert-validated particle picking annotations.

\paragraph{Baselines and backbones.}
We select representative and widely used baselines for cryo-EM image denoising, including classical signal processing approaches and SOTA learning-based methods. 
Specifically, we include classical \textbf{Gaussian} low-pass filter, \textbf{Topaz-Denoise}, and \textbf{DRACO}. 
These baselines cover both traditional filtering-based techniques and modern deep learning frameworks, providing a comprehensive comparison across different methodological paradigms.

\paragraph{Downstream cryo-EM processing.}
To assess the impact of denoising on structural reconstruction, we perform a standardized cryo-EM processing pipeline. 
Particle picking is conducted using the Topaz particle picker~\cite{bepler2019positive} integrated in cryoSPARC software~\cite{punjani2017cryosparc}, employing the default pretrained ResNet16 model (32 feature units) without additional fine-tuning. 
All subsequent cryo-EM data processing steps, including particle extraction, 2D classification, ab initio reconstruction, and refinement, are performed in cryoSPARC. 
This unified pipeline ensures a fair and reproducible evaluation of downstream performance.

\paragraph{Evaluation metrics.}
Since clean ground-truth micrographs are unavailable in modern cryo-EM, image-level metrics such as PSNR and SSIM are not applicable. 
We therefore evaluate denoising performance through downstream cryo-EM tasks. 
For particle picking, we use a distance-based detection metric, where a prediction is considered correct if it matches a ground-truth particle within a predefined distance threshold under one-to-one matching. 
We report precision, recall, and F1 under both micro- and macro-averaging. Micro scores are computed globally by aggregating counts across all micrographs, while macro scores are averaged over per-micrograph metrics.
For structural reconstruction, we report the resolution of the reconstructed 3D density map, estimated using the gold-standard Fourier Shell Correlation (FSC) 0.143 criterion.

\subsection{Overall Performance}

\subsubsection{Denoising Performance}
\begin{figure*}[t]
    \centering
    \setlength{\tabcolsep}{2pt}
    \renewcommand{\arraystretch}{1.0}

    \begin{tabular}{c c c c c c c}
        & \textbf{Original} & \textbf{Gaussian} & \textbf{DRACO} & \textbf{Topaz} & \textbf{DSM} & \textbf{TSM} \\

        \raisebox{2.25em}{\parbox[c][0.1\textwidth][c]{0.3cm}{\centering\rotatebox{90}{\textbf{10081}}}} &
        \includegraphics[width=0.14\textwidth]{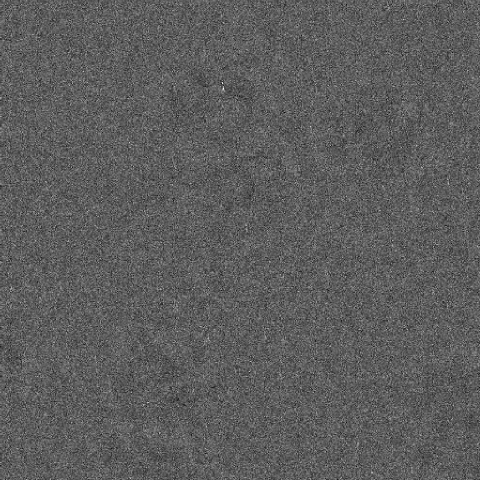} &
        \includegraphics[width=0.14\textwidth]{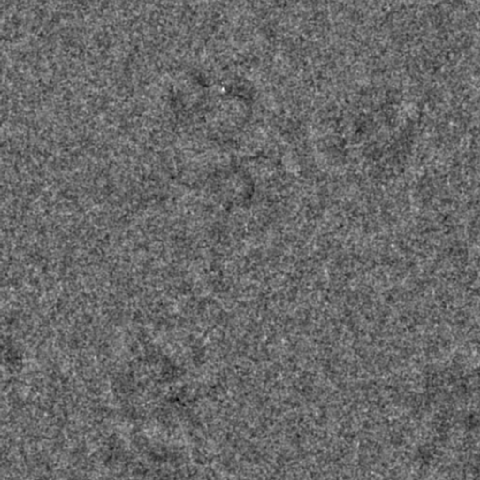} &
        \includegraphics[width=0.14\textwidth]{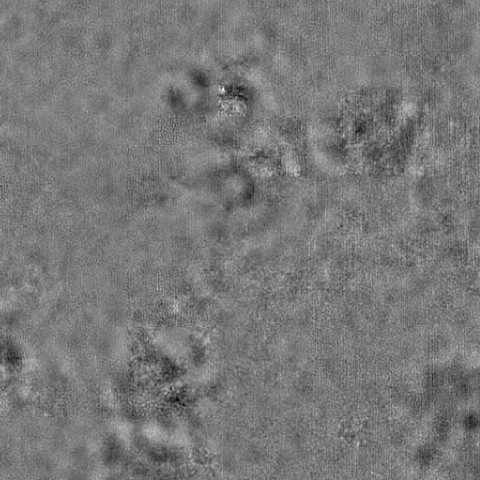} &
        \includegraphics[width=0.14\textwidth]{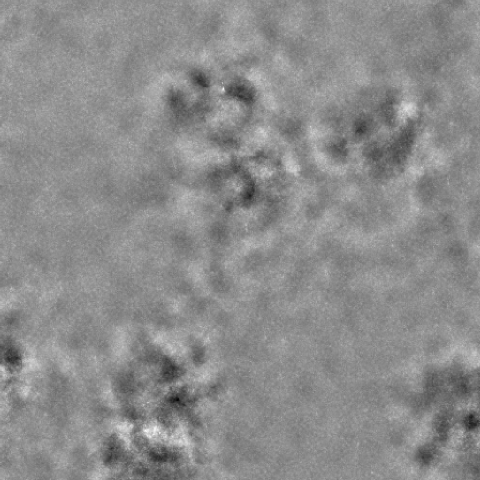} &
        \includegraphics[width=0.14\textwidth]{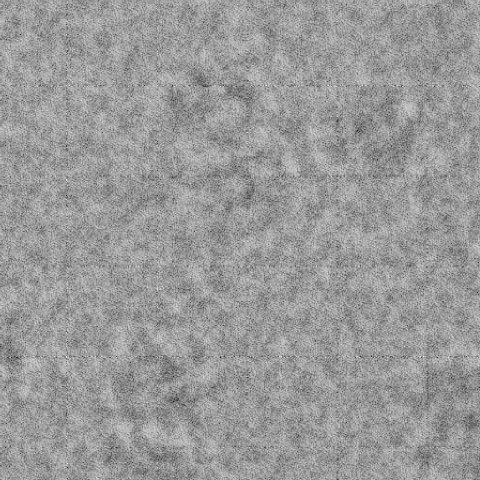} &
        \includegraphics[width=0.14\textwidth]{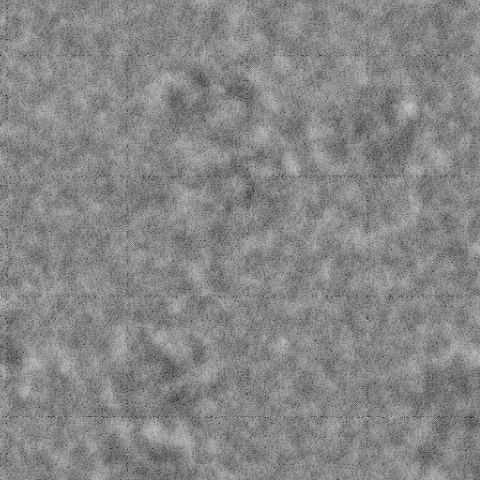} \\

        \raisebox{2.25em}{\parbox[c][0.1\textwidth][c]{0.3cm}{\centering\rotatebox{90}{\textbf{10289}}}} &
        \includegraphics[width=0.14\textwidth]{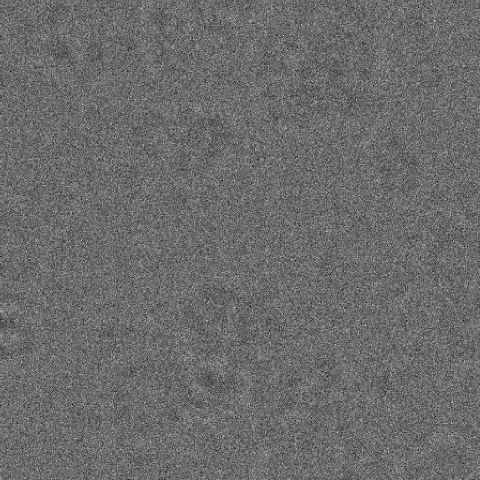} &
        \includegraphics[width=0.14\textwidth]{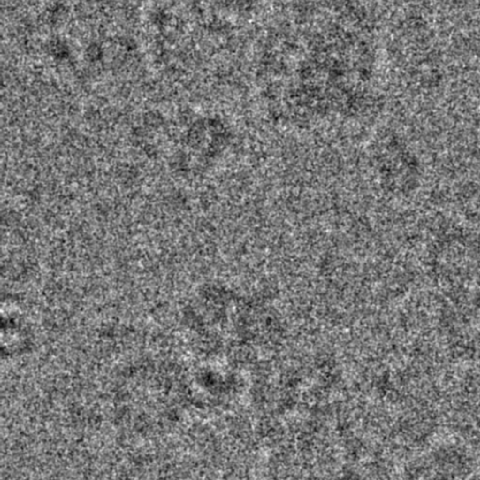} &
        \includegraphics[width=0.14\textwidth]{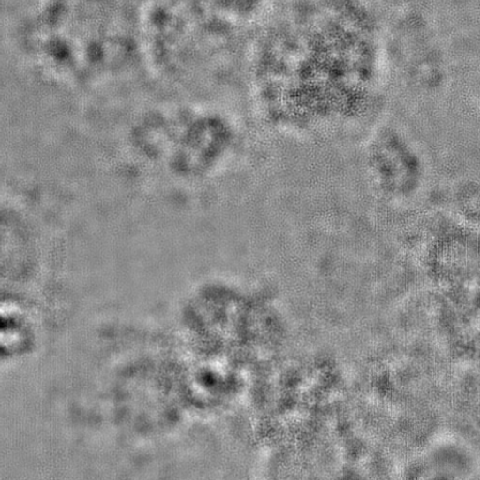} &
        \includegraphics[width=0.14\textwidth]{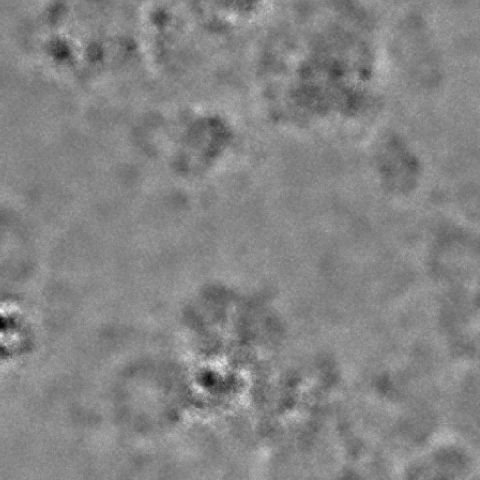} &
        \includegraphics[width=0.14\textwidth]{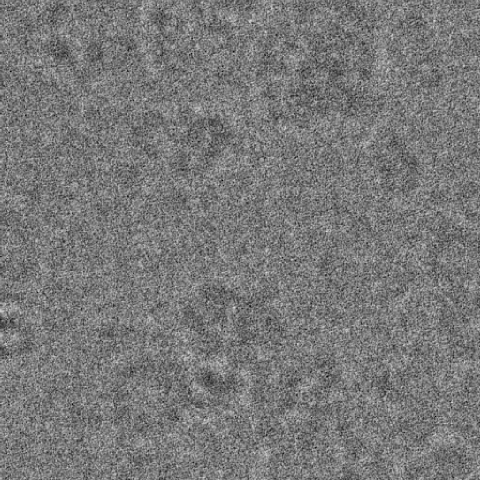} &
        \includegraphics[width=0.14\textwidth]{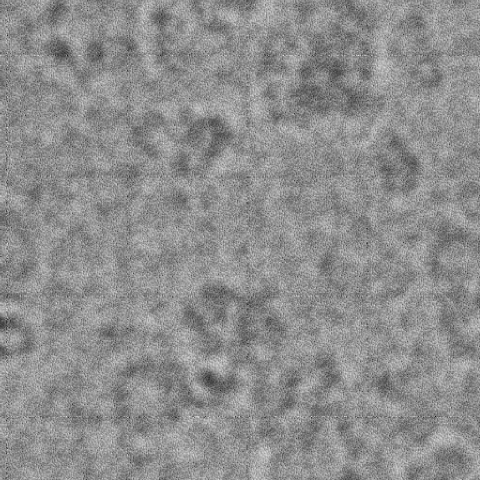} \\

        \raisebox{2.25em}{\parbox[c][0.1\textwidth][c]{0.3cm}{\centering\rotatebox{90}{\textbf{10291}}}} &
        \includegraphics[width=0.14\textwidth]{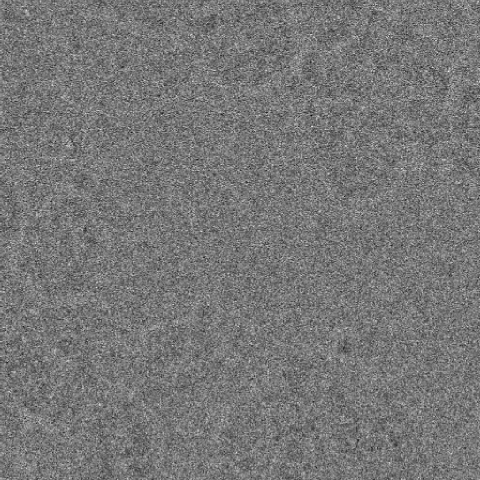} &
        \includegraphics[width=0.14\textwidth]{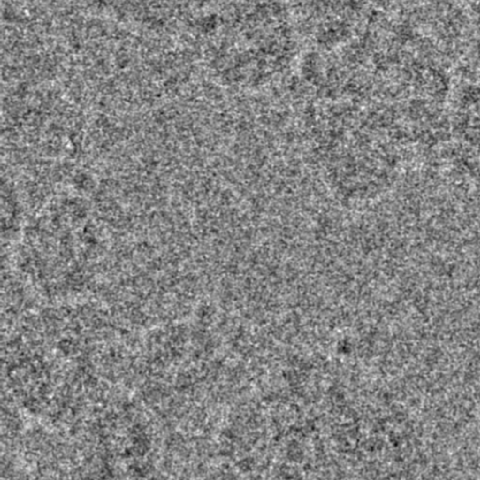} &
        \includegraphics[width=0.14\textwidth]{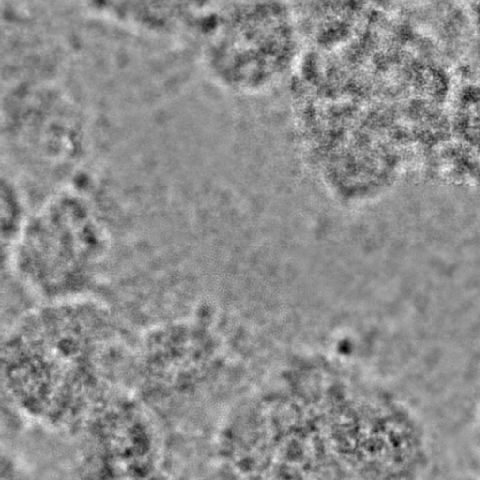} &
        \includegraphics[width=0.14\textwidth]{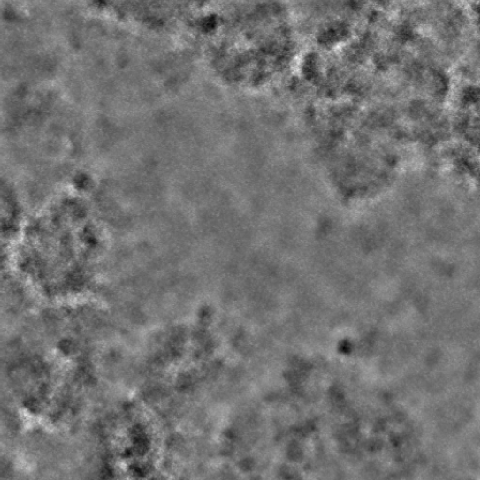} &
        \includegraphics[width=0.14\textwidth]{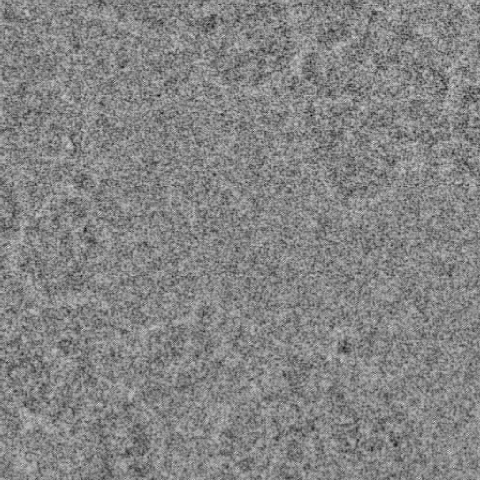} &
        \includegraphics[width=0.14\textwidth]{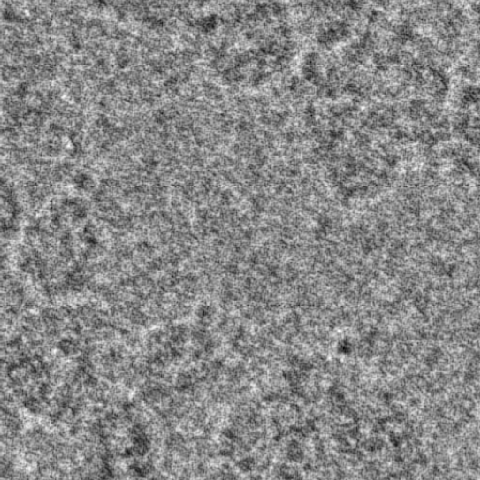} \\
    \end{tabular}

    \caption{Qualitative denoising comparisons on three cryo-EM datasets.}
    \label{fig:denoising_comparison}
\end{figure*}

The qualitative comparisons in Fig.~\ref{fig:denoising_comparison} suggest that effective cryo-EM denoising enhances particle-relevant structure rather than merely altering image appearance. Gaussian filtering provides only limited enhancement, leaving many weak particle patterns difficult to distinguish. DRACO and Topaz produce stronger responses, but these changes are accompanied by less stable local structure. By contrast, the score-based methods generate more particle-consistent patterns while better preserving the overall structure. TSM is particularly effective in this regard, yielding clearer particle visibility and more reliable cues for downstream particle picking. Such improved visual interpretability also facilitates manual inspection and particle annotation. These observations matches the following downstream picking results.

\subsubsection{Particle Picking Performance}

\begin{table}[t]
\centering
\small
\resizebox{0.9\textwidth}{!}{
\begin{tabular}{c|ccc|ccc|ccc}
\toprule
 & \multicolumn{3}{c}{EMPIAR-10291}
 & \multicolumn{3}{c}{EMPIAR-10289}
 & \multicolumn{3}{c}{EMPIAR-10081} \\
\cmidrule(lr){2-4} \cmidrule(lr){5-7} \cmidrule(lr){8-10}
Method & P & R & F1 & P & R & F1 & P & R & F1 \\
\midrule
Original & \textbf{0.617} & 0.646 & 0.631 
         & 0.352 & 0.287 & 0.316 
         & 0.546 & 0.738 & 0.627 \\
\midrule
Gaussian & 0.610 & 0.675 & 0.641 
         & 0.352 & 0.287 & 0.316 
         & 0.546 & 0.739 & 0.628 \\
\midrule
DRACO    & 0.542 & \textbf{0.723} & 0.620 
         & \textbf{0.384} & 0.351 & 0.367 
         & 0.383 & 0.456 & 0.417 \\
Topaz    & 0.539 & 0.710 & 0.613 
         & \textbf{0.384} & 0.350 & 0.366 
         & 0.389 & 0.459 & 0.421 \\
\midrule
DSM      & 0.601 & 0.694 & 0.644 
         & 0.360 & 0.304 & 0.330 
         & 0.567 & \textbf{0.755} & \textbf{0.648} \\
TSM      & 0.590 & 0.721 & \textbf{0.649} 
         & 0.382 & \textbf{0.358} & \textbf{0.370} 
         & \textbf{0.577} & 0.729 & 0.644 \\
\bottomrule
\end{tabular}
}
\caption{Micro-averaged particle picking performance comparison.}
\label{tab:particle_results}
\end{table}

\begin{figure}[t]
\centering

\begin{minipage}[t]{0.64\textwidth}
\centering
\small
\resizebox{\textwidth}{!}{
\begin{tabular}{c|c|c|c}
\toprule
Method 
& EMPIAR-10291 
& EMPIAR-10289
& EMPIAR-10081 \\
\midrule
Original 
& 0.623 $\pm$ 0.073 
& 0.313 $\pm$ 0.048 
& 0.610 $\pm$ 0.111 \\
\midrule
Gaussian 
& 0.633 $\pm$ 0.071 
& 0.313 $\pm$ 0.048 
& 0.610 $\pm$ 0.111 \\
\midrule
DRACO    
& 0.610 $\pm$ 0.082 
& 0.359 $\pm$ 0.062 
& 0.390 $\pm$ 0.132 \\
Topaz    
& 0.604 $\pm$ 0.081 
& 0.360 $\pm$ 0.055 
& 0.393 $\pm$ 0.136 \\
\midrule
DSM      
& 0.636 $\pm$ 0.072 
& 0.325 $\pm$ 0.047 
& \textbf{0.626 $\pm$ 0.118} \\
TSM 
& \textbf{0.640 $\pm$ 0.076} 
& \textbf{0.364 $\pm$ 0.048} 
& 0.623 $\pm$ 0.115 \\
\bottomrule
\end{tabular}
}
\captionof{table}{Macro-averaged F1 score (mean $\pm$ std across micrographs) under distance-based evaluation.}
\label{tab:macro_all}
\end{minipage}
\begin{minipage}[t]{0.35\textwidth}
\centering
\setlength{\tabcolsep}{2pt}

\begin{subfigure}[b]{0.4\linewidth}
\includegraphics[width=\linewidth]{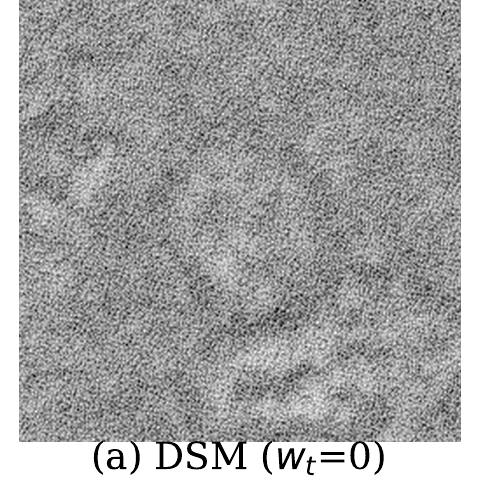}
\end{subfigure}
\begin{subfigure}[b]{0.4\linewidth}
\includegraphics[width=\linewidth]{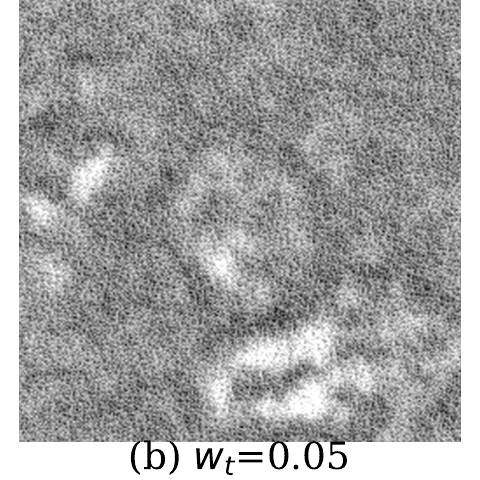}
\end{subfigure}

\begin{subfigure}[b]{0.4\linewidth}
\includegraphics[width=\linewidth]{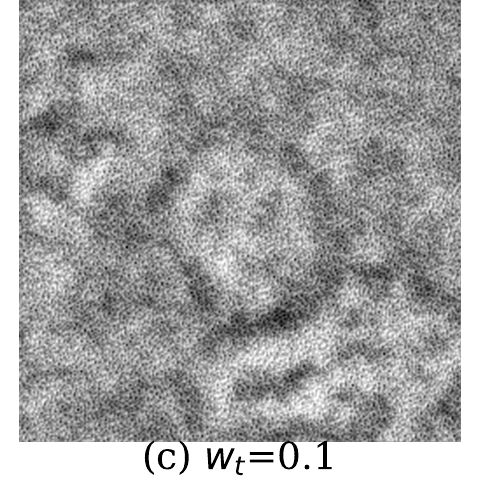}
\end{subfigure}
\begin{subfigure}[b]{0.4\linewidth}
\includegraphics[width=\linewidth]{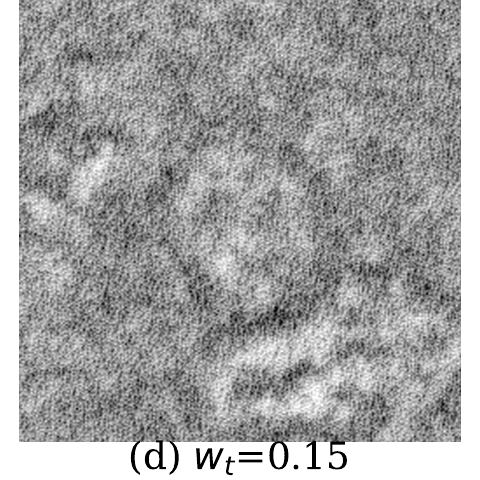}
\end{subfigure}

\caption{Ablation of the target weight $w_t$ on EMPIAR-10289.}
\label{fig:ablation_w_t}

\end{minipage}

\end{figure}

Tables~\ref{tab:particle_results} and \ref{tab:macro_all} report downstream particle picking performance under the distance-based (32px) evaluation protocol using both micro- and macro-averaged metrics. 
Compared with DRACO and Topaz, our score-based methods consistently achieve stronger overall performance across datasets, indicating that effective cryo-EM denoising should improve particle-relevant structure rather than merely amplifying particle-like responses.

Within the score-based family, TSM achieves the best F1 on EMPIAR-10291 and EMPIAR-10289 under both micro and macro evaluation, while DSM performs best on EMPIAR-10081 and TSM remains highly competitive. 
This difference is consistent with our theoretical motivation: target-guided supervision provides a more stable training signal in the low-noise regime and is particularly beneficial when particle localization depends on subtle structural cues. 
On EMPIAR-10081, DSM achieves the highest F1 while TSM attains the highest precision, suggesting that both score-based variants improve downstream picking but favor different precision--recall operating points.

\subsubsection{3D Reconstruction Performance}


\begin{figure*}[t]
\centering
\setlength{\tabcolsep}{1pt}
\renewcommand{\arraystretch}{1.0}

{\bfseries EMPIAR-10291}

\begin{tabular}{>{\centering\arraybackslash}m{0.13\textwidth}
                >{\centering\arraybackslash}m{0.13\textwidth}
                >{\centering\arraybackslash}m{0.13\textwidth}
                >{\centering\arraybackslash}m{0.13\textwidth}
                >{\centering\arraybackslash}m{0.13\textwidth}
                >{\centering\arraybackslash}m{0.13\textwidth}
                >{\centering\arraybackslash}m{0.13\textwidth}}
\includegraphics[width=0.13\textwidth]{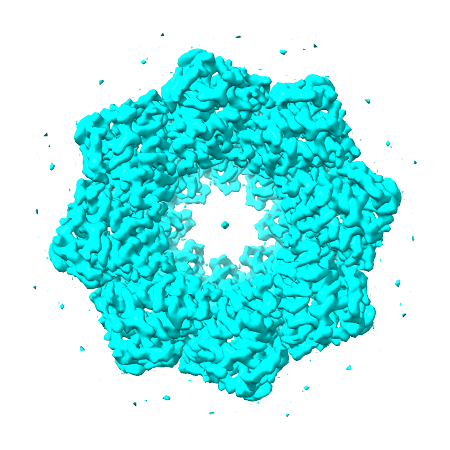} &
\includegraphics[width=0.13\textwidth]{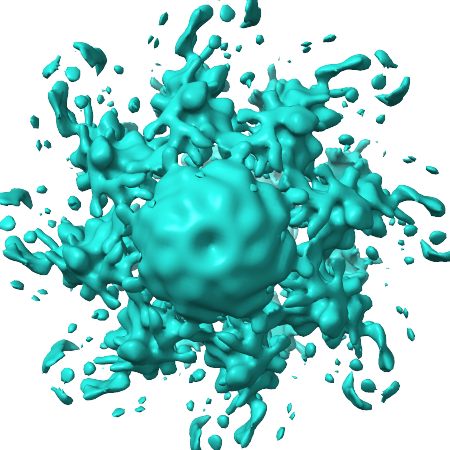} &
\includegraphics[width=0.13\textwidth]{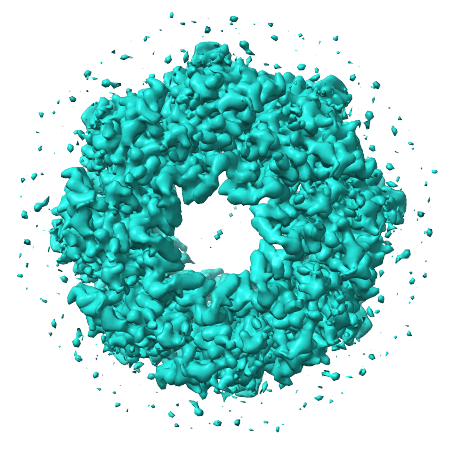} &
\includegraphics[width=0.13\textwidth]{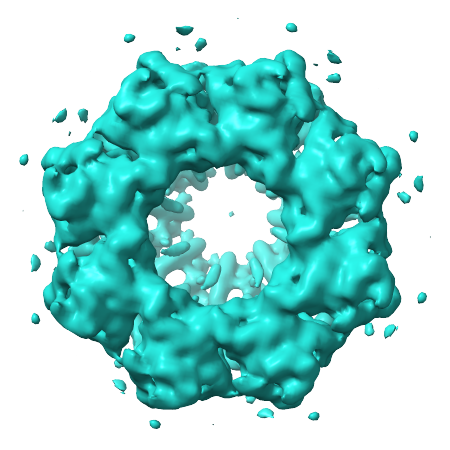} &
\includegraphics[width=0.13\textwidth]{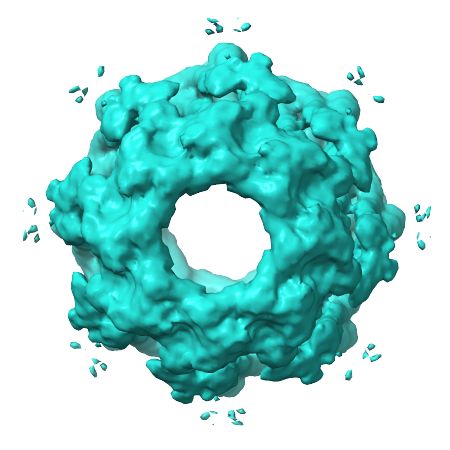} &
\includegraphics[width=0.13\textwidth]{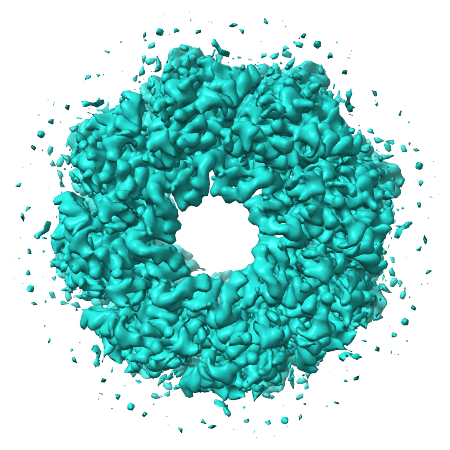} &
\includegraphics[width=0.13\textwidth]{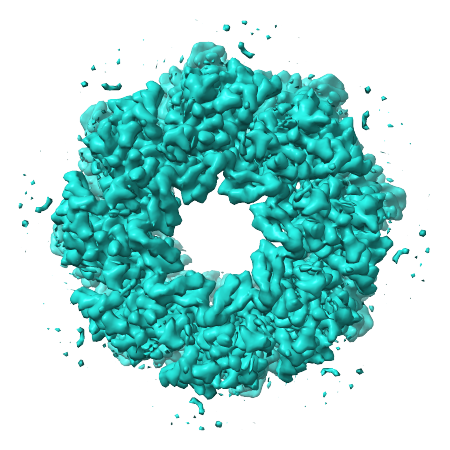} \\

GT Map (3.6\AA) &
Original (6.61\AA) &
Gaussian (3.85\AA) &
DRACO (6.08\AA) &
Topaz (4.68\AA) &
DSM (3.78\AA) &
TSM (3.75\AA) \\
\end{tabular}

{\bfseries EMPIAR-10081}

\begin{tabular}{>{\centering\arraybackslash}m{0.13\textwidth}
                >{\centering\arraybackslash}m{0.13\textwidth}
                >{\centering\arraybackslash}m{0.13\textwidth}
                >{\centering\arraybackslash}m{0.13\textwidth}
                >{\centering\arraybackslash}m{0.13\textwidth}
                >{\centering\arraybackslash}m{0.13\textwidth}
                >{\centering\arraybackslash}m{0.13\textwidth}}
\includegraphics[width=0.13\textwidth]{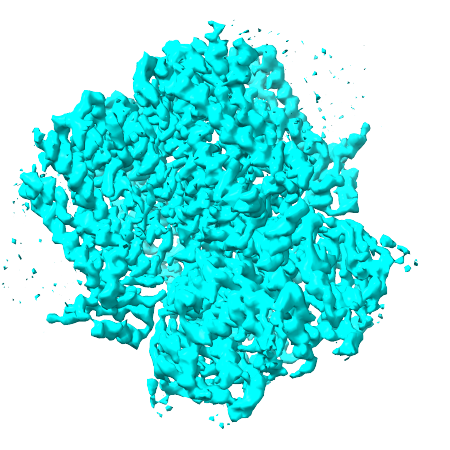} &
\includegraphics[width=0.13\textwidth]{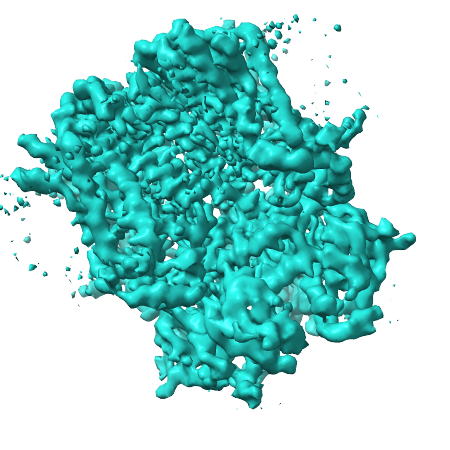} &
\includegraphics[width=0.13\textwidth]{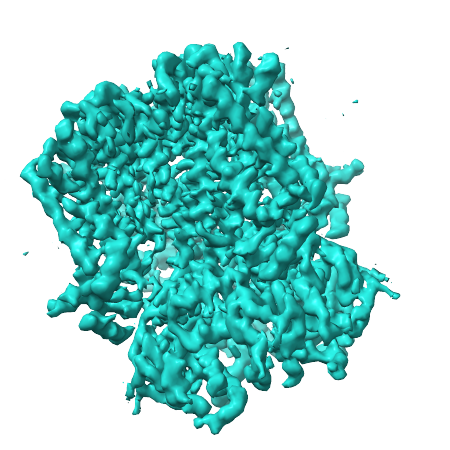} &
\includegraphics[width=0.13\textwidth]{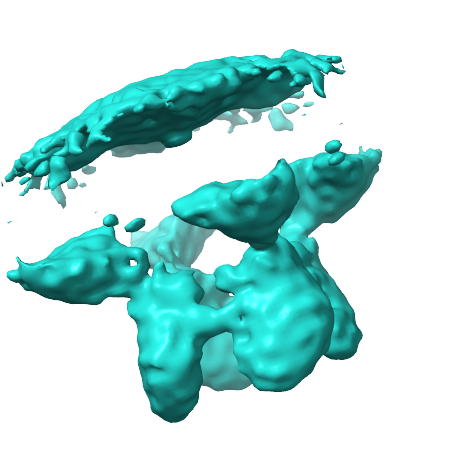} &
\includegraphics[width=0.13\textwidth]{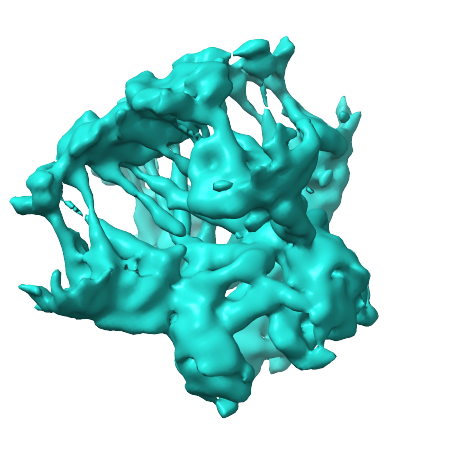} &
\includegraphics[width=0.13\textwidth]{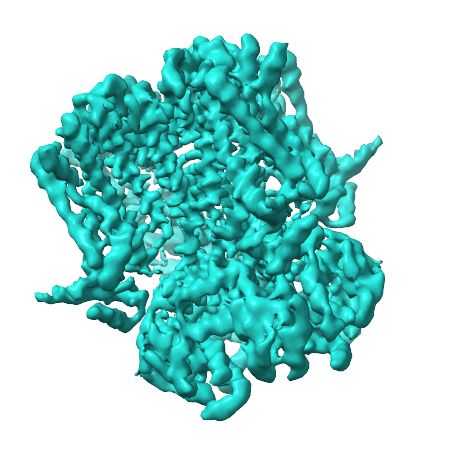} &
\includegraphics[width=0.13\textwidth]{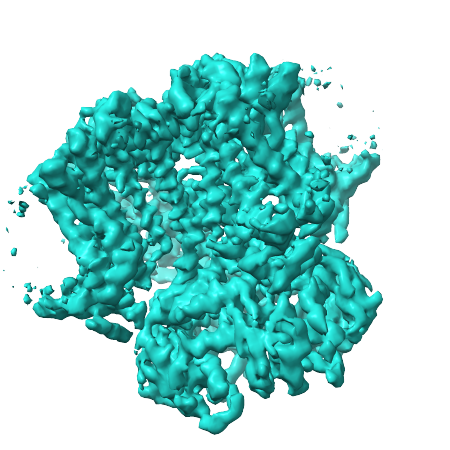} \\

GT Map (3.5\AA) &
Original (4.04\AA) &
Gaussian (4.05\AA) &
DRACO (6.39\AA) &
Topaz (7.35\AA) &
DSM (4.13\AA) &
TSM (4.10\AA) \\
\end{tabular}

\caption{3D reconstruction visualizations on EMPIAR-10291 and EMPIAR-10081. Each row shows the ground truth (GT) and reconstructions from the original input and five compared methods. Reported values indicate reconstruction resolution (lower is better).}
\label{fig:3d_recon_vis}
\end{figure*}

\begin{figure}[t]
    \centering
    \begin{subfigure}[t]{0.49\linewidth}
        \centering
        \includegraphics[width=\linewidth]{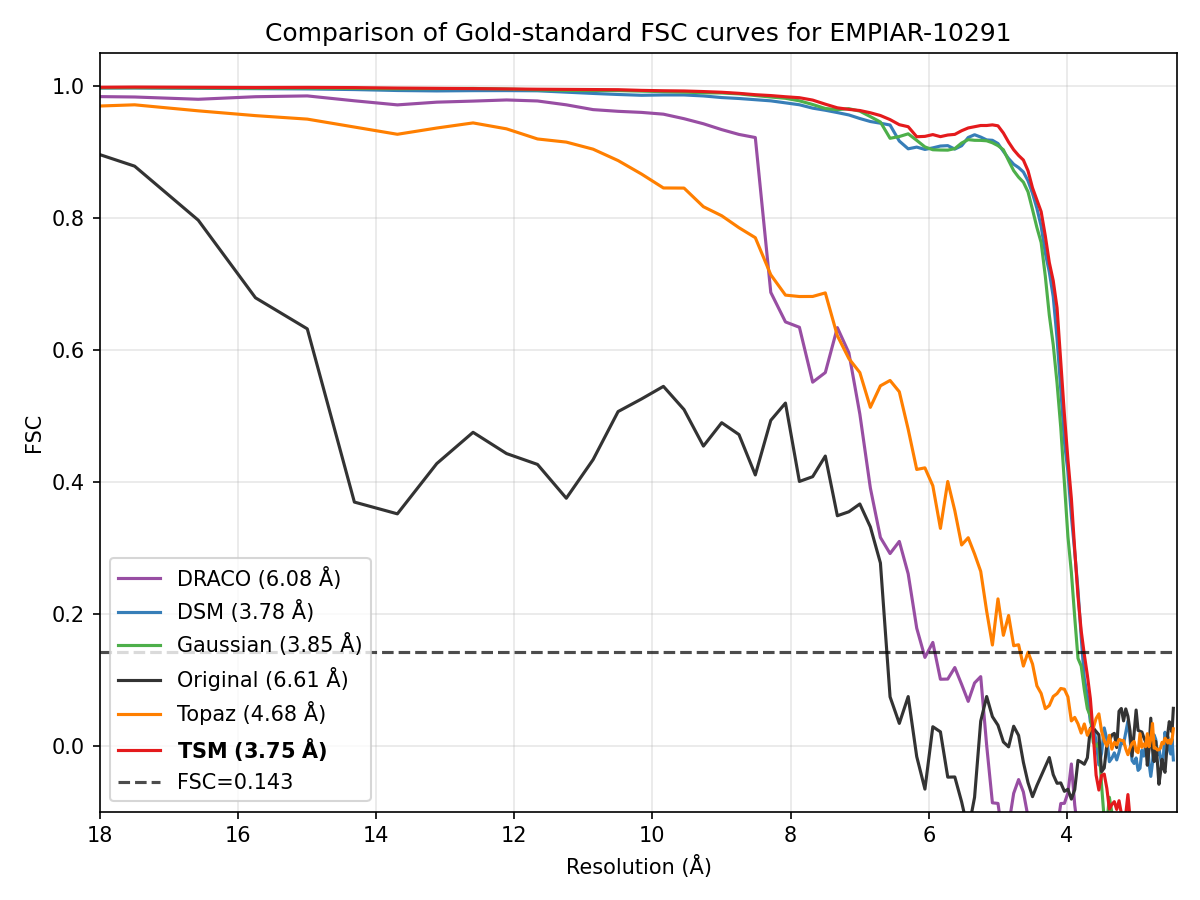}
        \caption{EMPIAR-10291}
        \label{fig:fsc_10291}
    \end{subfigure}
    \hfill
    \begin{subfigure}[t]{0.49\linewidth}
        \centering
        \includegraphics[width=\linewidth]{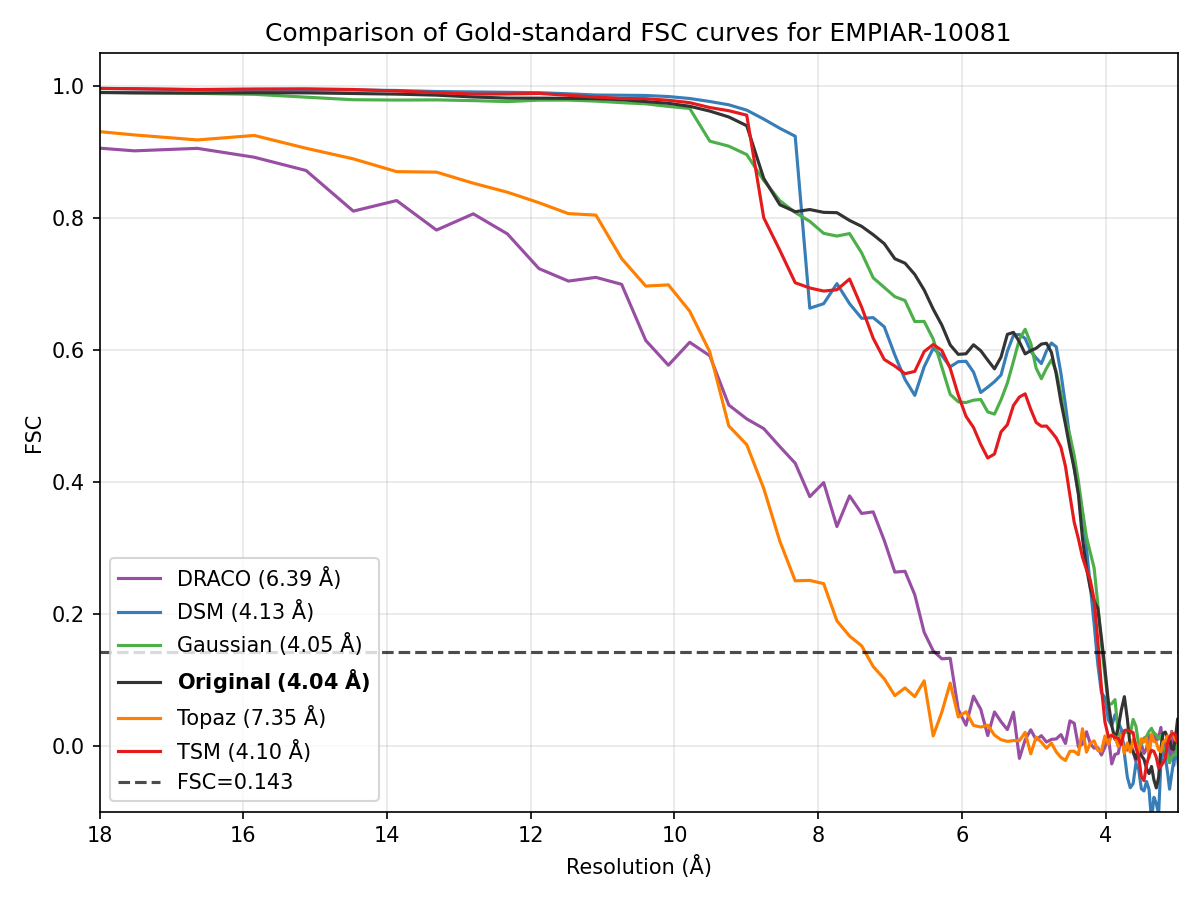}
        \caption{EMPIAR-10081}
        \label{fig:fsc_10081}
    \end{subfigure}
    \caption{FSC curves of 3D reconstructions comparison on EMPIAR-10291 and EMPIAR-10081. Resolution is determined at FSC $0.143$; a right-shifted curve (higher spatial frequency at the threshold) indicates better resolution.}
    \label{fig:fsc_two_datasets}
\end{figure}

The FSC curves for 3D reconstructions on EMPIAR-10291 and EMPIAR-10081 are shown in Fig.~\ref{fig:fsc_two_datasets}, with the corresponding density maps visualized in Fig.~\ref{fig:3d_recon_vis}. Reconstruction resolution is evaluated using the gold-standard FSC $0.143$ criterion, where a lower resolution value indicates better structural recovery. We do not include EMPIAR-10289 here because it targets the same macromolecular complex as EMPIAR-10291 but was collected under different microscope settings and experimental conditions. Overall, DSM and TSM achieve substantially better reconstruction quality than the other learned denoising baselines, particularly DRACO and Topaz. This trend is also consistent with the reconstructed density maps. 

On EMPIAR-10081, DRACO leads to severe structural degradation, with substantial missing regions, while DSM and TSM preserve both the overall morphology and local structural details much more faithfully, and their reconstructions remain visually much closer to the GT maps. Although Gaussian filtering can also produce reasonable reconstructions, achieving such performance requires careful parameter tuning, which makes it less reliable in practice. We also note that small differences in the reported resolution values do not necessarily imply substantial differences in reconstruction quality. When the values are close (i.e., difference within $0.1$\AA), the practical difference is often limited and should be interpreted together with the FSC curves and the visualized density maps.

These observations are also aligned with the particle-picking results. On EMPIAR-10291, although the original micrographs already achieve reasonable overall picking performance, the remaining false positives in the picked particle set can still degrade downstream reconstruction and are consistent with the artifacts observed in the reconstructed map. In contrast, DSM and TSM yield cleaner reconstructions, suggesting that denoising is more effective when it improves particle detectability while reducing false positives and preserving particle-level structural information for downstream alignment and averaging.

\subsection{Ablation Study \& Sensitivity Analysis}
\label{ablation_sensitivity}
\begin{figure}[t]
\centering

\begin{subfigure}[b]{0.55\textwidth}
    \centering
    \includegraphics[width=0.3\linewidth]{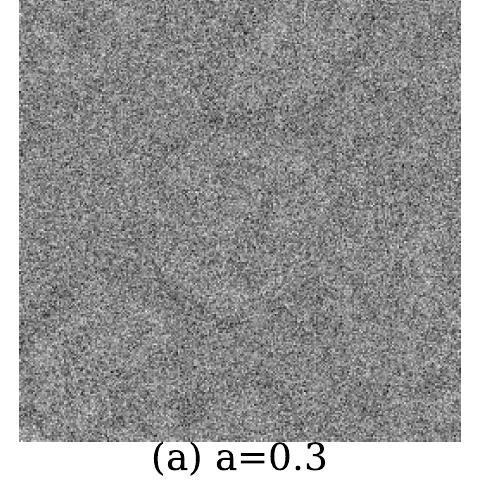}
    \includegraphics[width=0.3\linewidth]{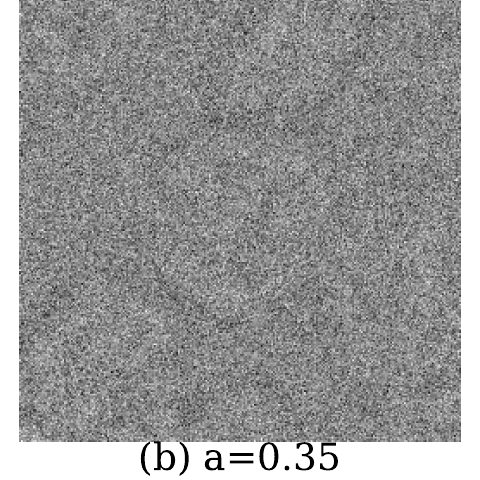}
    \includegraphics[width=0.3\linewidth]{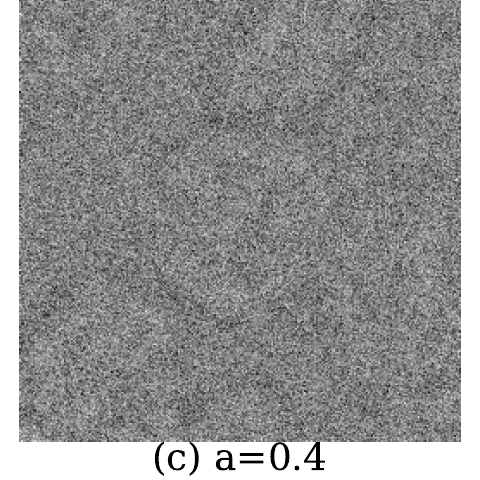}
    \includegraphics[width=0.3\linewidth]{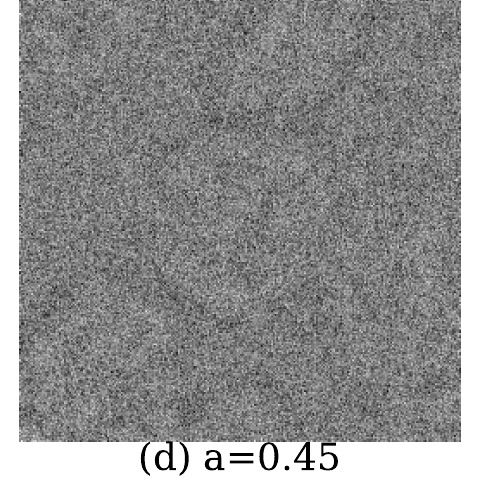}
    \includegraphics[width=0.3\linewidth]{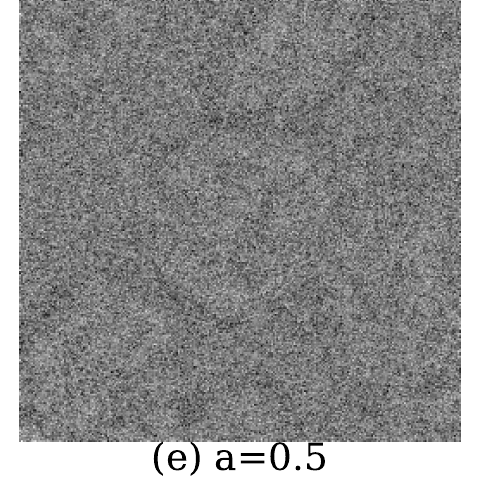}
    \includegraphics[width=0.3\linewidth]{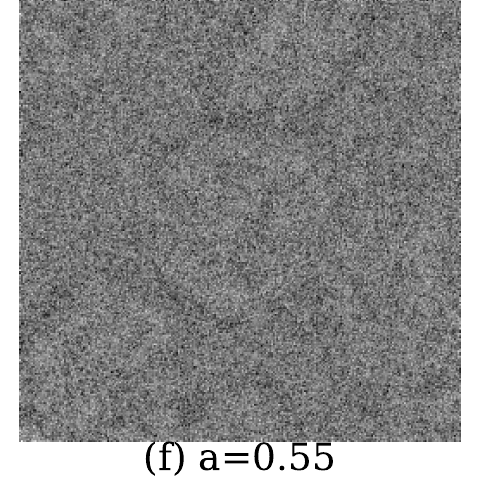}
    \caption{Sensitivity analysis w.r.t. the noise modeling parameter $a$}
\end{subfigure}
\hfill
\begin{subfigure}[b]{0.44\textwidth}
    \centering
    \includegraphics[width=\linewidth]{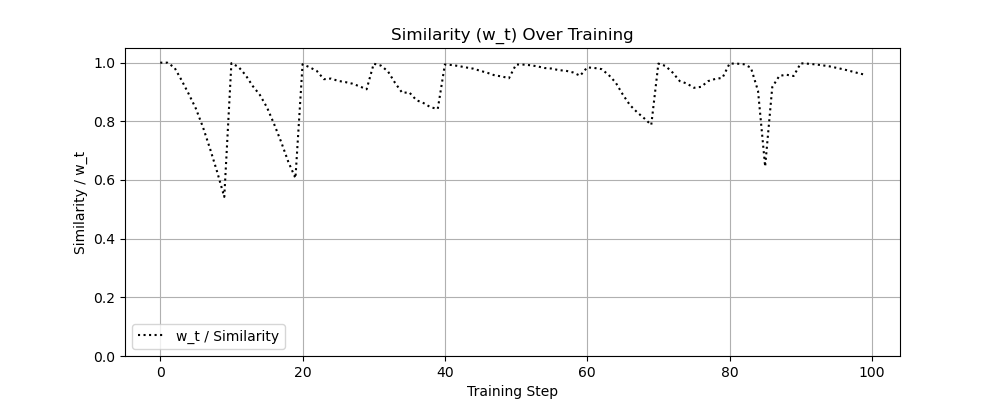}
    \includegraphics[width=\linewidth]{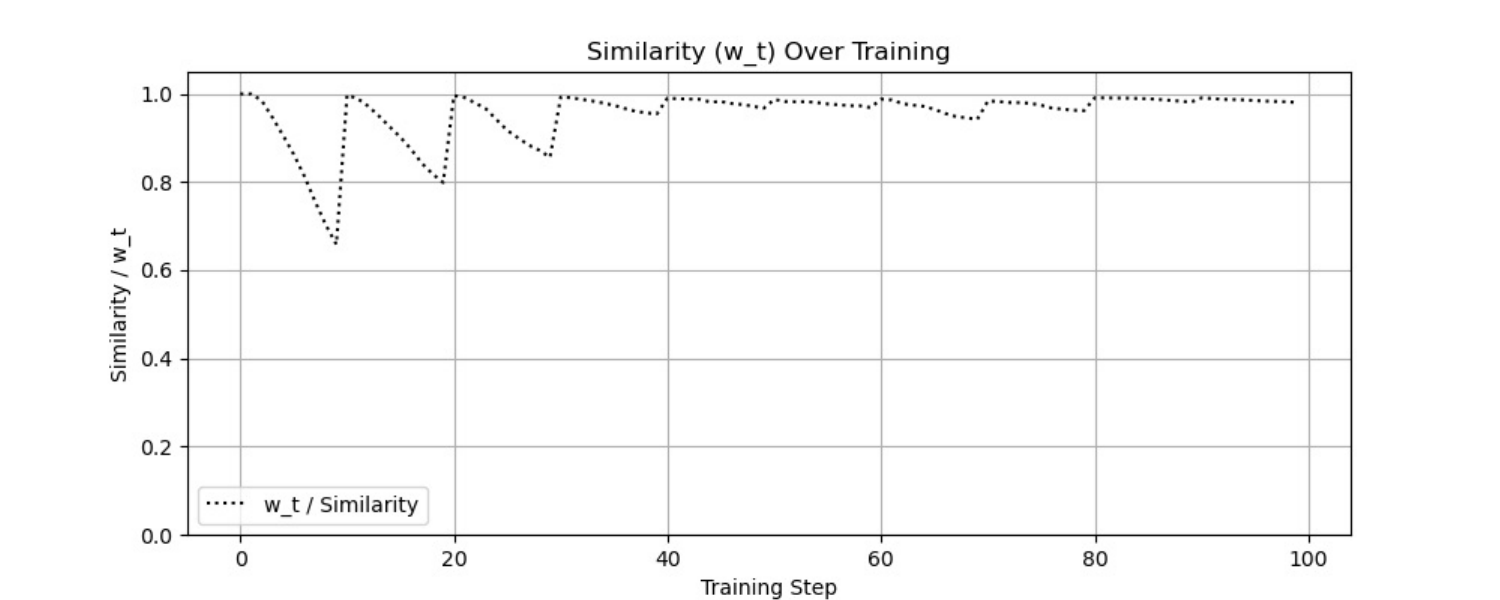}
    \caption{Effect of annealed learning (top: w/o annealing; bottom: w/ annealing).}
\end{subfigure}
\caption{Left: sensitivity analysis under different noise simulation parameters $a$. 
Right: comparison of training w/ and w/o annealed learning.}
\label{fig:sensitivity_vs_anneal}
\end{figure}

\paragraph{\textbf{Target weight $w_t$.}}
We study the effect of the target weight $w_t$ on EMPIAR-10289. Fig.~\ref{fig:ablation_w_t} compares DSM-only ($w_t=0$) with $w_t\in\{0.05,0.1,0.15\}$. Without target guidance (Fig.~\ref{fig:ablation_w_t}a), DSM mainly reduces noise while leaving particle structures poorly defined. Enabling TSM (Figs.~\ref{fig:ablation_w_t}b--d) consistently improves structural clarity across different values of $w_t$, producing sharper particle boundaries and reduced background responses.
The improvement remains stable over a wide range of $w_t$, indicating low sensitivity to precise parameter tuning once target guidance is introduced. When $w_t$ becomes excessively large (Fig.~\ref{fig:ablation_w_t}d), mild over-regularization begins to suppress fine structural variations. Overall, moderate values of $w_t$ achieve the best balance between structural guidance and preservation of fine details.

\paragraph{\textbf{Global guidance coefficient $\lambda(t)$.}}
We evaluate the effect of the global guidance coefficient $\lambda(t)$ on EMPIAR-10291, which controls the contribution of TSM during training. Fig.~\ref{fig:sensitivity_vs_anneal}b compares training without annealing (top) and with annealed learning (bottom), where the similarity weight $w_t$ between a fixed noisy input and its plausible targets is tracked across training. The encoder is updated every 10 epochs. Without annealing, TSM is applied from the beginning, when the encoder is still unstable; consequently, the projected noisy–target features are unreliable, and the similarity $w_t$ exhibits noticeable oscillations after each feature update. These fluctuations indicate that early similarity estimates are not trustworthy, leading to unstable target guidance. In contrast, with annealed learning we set $\lambda(t)=0$ in the first phase to optimize only DSM, allowing the encoder to learn stable feature representations before introducing target guidance. The coefficient $\lambda(t)$ is then increased linearly in the middle phase and fixed to $\lambda(t)=1$ in the final phase. As shown in Fig.~\ref{fig:sensitivity_vs_anneal}b (bottom), this schedule significantly stabilizes the similarity curve, producing consistently high and less oscillatory $w_t$ values. Overall, annealed learning improves the reliability of similarity estimation and yields more stable optimization dynamics by delaying target guidance until the encoder becomes sufficiently well-structured.

\paragraph{\textbf{Sensitivity to noise modeling coefficient $\sigma_a$.}}
Fig.~\ref{fig:sensitivity_vs_anneal}a evaluates the sensitivity to the noise modeling coefficient $a$, which controls the noise scale $\sigma_a$. Qualitatively, the method produces consistent visual results across a wide range of $a$, and a moderate range ($a\in[0.4,0.5]$) offers a favorable trade-off between noise suppression and structure detail preservation. Smaller $a$ leaves more residual noise, whereas larger $a$ slightly oversmooths fine structural details. Overall, the visual quality shows limited sensitivity to the choice of $\sigma_a$.
\section{Conclusion}

In conclusion, we introduced a target-guided score-based denoising framework for cryo-EM to address extreme noise and weak particle signals. Our design leverages score-based learning to preserve particle-relevant structure without overly suppressing non-particle regions, and further improves particle-region accuracy via target-informed guidance. Extensive experiments across multiple cryo-EM datasets show consistent downstream benefits: our method improves particle picking over classical signal-processing baselines, remains competitive with strong deep learning denoisers, and supports high-quality 3D reconstruction, as reflected by FSC trends and cleaner density maps.

Looking ahead, improving target construction under dataset-specific and heterogeneous noise conditions remains an important direction. While similarity-guided annealing provides stable guidance during training, richer target representations and more accurate noise modeling could further improve robustness on particularly challenging micrographs. Beyond cryo-EM, the proposed target-informed score-matching formulation is general and could extend to other low-SNR scientific imaging problems, including cryo-electron tomography, fluorescence microscopy, and other physics-driven inverse problems where structural preservation is critical.

%
%
\section*{Acknowledgement}
This work was supported in part by U.S. NIH grant R35GM158094.
\bibliographystyle{splncs04}
\bibliography{main,Supplementary_material}

\appendix  
\clearpage
\input{Supplementary_material}

\end{document}

%% file: Supplementary_material.tex
\section{Derivations and Theoretical Details}
\subsection{Posterior Score Identity for the Corruption Model}
\label{sec:posterior_score_identity}

We recall a classical identity relating the score of the noisy observation
distribution to the conditional score of the corruption model in Sec.~\ref{sec:problem_def}.
For completeness, we include a short derivation here,
following the argument presented in Appendix~A of
\cite{scorebased}, without any claim for originality.

Let $X\sim p_X$ and $Y\sim p_Y$ denote the noisy and clean patch-level
random variables respectively.
Let the corruption model be defined by the conditional distribution
$p_{X|Y}(x|y)$.
We define the conditional score of the corruption model as
\[
s_{X|Y}(x|y) := \nabla_x \log p_{X|Y}(x|y).
\]

\paragraph{Proposition.}
The score of the noisy distribution $p_X$ can be written as
\begin{equation}
\nabla_x \log p_X(x)
=
\int p_{Y|X}(y|x)\, s_{X|Y}(x|y)\, dy .
\label{eq:posterior_score_identity}
\end{equation}

\paragraph{Proof.}
The marginal density of the noisy observation is obtained by
marginalizing the latent clean variable:
\begin{equation}
p_X(x)
=
\int p_{X|Y}(x|y)\, p_Y(y)\, dy .
\end{equation}

Taking the gradient with respect to $x$ gives
\begin{equation}
\nabla_x p_X(x)
=
\int \nabla_x p_{X|Y}(x|y)\, p_Y(y)\, dy .
\end{equation}

Using the identity
\(
\nabla_x p_{X|Y}(x|y)
=
p_{X|Y}(x|y)\, \nabla_x \log p_{X|Y}(x|y)
\),
we obtain
\begin{equation}
\nabla_x p_X(x)
=
\int
p_{X|Y}(x|y)\,
s_{X|Y}(x|y)\,
p_Y(y)\, dy .
\end{equation}

Dividing both sides by $p_X(x)$ yields
\begin{align}
\nabla_x \log p_X(x)
&=
\int
\frac{p_{X|Y}(x|y)\, p_Y(y)}{p_X(x)}
\, s_{X|Y}(x|y)\, dy \\
&=
\int
p_{Y|X}(y|x)\,
s_{X|Y}(x|y)\, dy ,
\end{align}
where we used Bayes' rule
\(
p_{Y|X}(y|x)
=
\frac{p_{X|Y}(x|y)p_Y(y)}{p_X(x)}.
\)

This establishes Eq.~\eqref{eq:posterior_score_identity}, which forms the basis of denoising score matching
\cite{scorebased}.

\subsection{Preconditioning the TSM objective}
\label{sec:preconditioning_training}

This section derives the rescaled objective used in
Sec.~\ref{sec:target_score_supervision}.
Our derivation follows the preconditioning analysis of
\cite{karras2022elucidatingdesignspacediffusionbased}
and adapts it to the target-score supervision in our framework.

\paragraph{Starting objective.}

The target-score supervision in
Sec.~\ref{sec:target_score_supervision}
trains a noisy score network
\(s_\theta(x)\)
to match the surrogate clean score
\[
\nabla_y \log p_Y^{\text{sur}}(y)=-\frac{y}{\sigma^2}.
\]

Given a training pair
\((x,\tilde{y}_{\mathrm{tsm}}(x))\),
the ideal regression objective is

\begin{equation}
L(\theta)
=
\mathbb{E}_x
\left[
\left\|
s_\theta(x)
+
\frac{\tilde{y}_{\mathrm{tsm}}(x)}{\sigma^2}
\right\|^2
\right].
\label{eq:tsm_base_loss}
\end{equation}

However, directly optimizing
\eqref{eq:tsm_base_loss}
leads to a mismatch of scale across noise levels,
since the magnitude of the optimal score depends on
the corruption variance \(\sigma_a^2\).

\paragraph{Network parameterization.}

Following the AR-DAE formulation
\cite{lim2020ardae}, the score network $s_\theta$ is implemented using
an internal neural network $F_\theta$. To improve conditioning across different noise levels,
we adopt the linear preconditioning scheme introduced in
\cite{karras2022elucidatingdesignspacediffusionbased}.
Under this scheme the score network is written as

\begin{equation}
s_\theta(x)
=
c_o
F_\theta(\sigma_a+c_i x)
+
c_s x ,
\end{equation}

where \(c_i,c_o,c_s\) are scalar coefficients that normalize the input,
network output, and skip connection respectively.

Substituting this parameterization into
\eqref{eq:tsm_base_loss}
and introducing a loss weight \(\lambda\)
yields the preconditioned objective

\begin{equation}
L(\theta)
=
\lambda
\,
\mathbb{E}_x
\left[
\left\|
c_o F_\theta(\sigma_a,c_i x)
+
c_s x
+
\frac{\tilde{y}_{\mathrm{tsm}}(x)}{\sigma^2}
\right\|^2
\right].
\label{eq:precond_general}
\end{equation}

The coefficients are chosen so that

\begin{itemize}

\item the input to \(F_\theta\) has unit variance,
\item the regression target has unit variance,
\item the effective loss weight \(\lambda c_o^2\) equals one,
\item the skip connection minimizes amplification of network errors.
\end{itemize}

\paragraph{Coefficient selection.}

Under the additive corruption model

\[
X = Y + N, \qquad
N\sim\mathcal{N}(0,\sigma_a^2 I),
\]

and the surrogate clean distribution

\[
Y\sim\mathcal{N}(0,\sigma^2 I),
\]

the variance of the noisy observation is

\begin{equation}
\mathrm{Var}(X)=\sigma^2+\sigma_a^2.
\end{equation}

Condition (i) therefore gives

\begin{equation}
c_i = \frac{1}{\sqrt{\sigma^2+\sigma_a^2}}.
\end{equation}

Minimizing output amplification yields the skip coefficient

\begin{equation}
c_s = -\frac{1}{\sigma^2+\sigma_a^2}.
\end{equation}

Using conditions (ii)–(iii) together with the derivation in
\cite{karras2022elucidatingdesignspacediffusionbased}
gives

\begin{align}
c_o &= -\frac{\sigma_a}{\sigma\sqrt{\sigma^2+\sigma_a^2}}, \\
\lambda &= \frac{\sigma^2\sigma_a^2+\sigma^4}{\sigma^2}.
\end{align}

\paragraph{Resulting preconditioned objective.}

Substituting these coefficients into
\eqref{eq:precond_general}
and rewriting the expression
in terms of the score network $s_\theta$ yields the final loss
used in the main paper.

\begin{align}
\ell_{\text{TSM}}(x;\theta)
&=
\frac{\sigma^2\sigma_a^2+\sigma^4}{\sigma^2}
\Bigg\|
\frac{\sigma_a}{\sigma\sqrt{\sigma^2+\sigma_a^2}}
s_\theta
\!\left(
\frac{x}{\sqrt{\sigma^2+\sigma_a^2}}+\sigma_a
\right)
+
\frac{x}{\sigma^2+\sigma_a^2}
-
\frac{\tilde{y}_{\mathrm{tsm}}(x)}{\sigma^2}
\Bigg\|_2^2 .
\end{align}

This preconditioning preserves the target-score supervision while
ensuring that the effective training signal remains well-conditioned
across different noise levels.

\section{Dataset Description}

\begin{table}[t]
\centering
\resizebox{\textwidth}{!}{
\begin{tabular}{c l c c c c}
\toprule
EMPIAR ID & Protein Type & \# Micrographs & Image Size & Diameter (px) & Diameter (\AA) \\
\midrule
10081 & HCN1 ion channel & 300 & $(3710, 3838)$ & 154 & 200 \\
10289 & innexin-6 hemichannels & 300 & $(3710, 3838)$ & 162 & 200 \\
10291 & innexin-6 hemichannels & 300 & $(3710, 3838)$ & 130 & 160 \\
\bottomrule
\end{tabular}
}
\caption{Summary of the EMPIAR datasets used in our experiments.}
\label{tab:dataset_summary}
\end{table}

We evaluate our method on three publicly available cryo-EM datasets from the EMPIAR repository. The selected datasets cover different particle types and sizes. We include both EMPIAR-10289 and EMPIAR-10291 because, although they are both derived from the same innexin-6 hemichannel study, they correspond to different structural settings and were acquired under different experimental conditions. In particular, the original study compares wild-type INX-6 in nanodiscs, wild-type INX-6 in detergent, and an N-terminal deletion mutant in nanodiscs, highlighting differences in N-terminal rearrangement and lipid environment. These related but distinct datasets provide a useful testbed for evaluating whether different denoising methods improve downstream particle picking and 3D reconstruction consistently across structural and imaging variations. Table~\ref{tab:dataset_summary} summarizes the selected datasets.

\section{Implementation Details}

\subsection{Patch-Wise Training Settings}

We implement our method using a diffusion-style U-Net backbone based on \texttt{GuidedDDPMPlainUNet} \cite{ho2020ddpm, dhariwal2021diffusion}. The network processes single-channel cryo-EM micrographs and consists of multi-scale residual blocks with base width 128 and channel multipliers $(1,2,2,4)$ to capture structural features across spatial scales. 

During training, we randomly sample $256\times256$ patches from micrographs, extracting 32 patches per micrograph per epoch. We adopt a GaussianMap noise model that combines Poisson and Gaussian noise with parameters $\alpha=50$ and $\sigma=0.05$, together with auxiliary noise levels $\{0.2,0.1,0.05,0.01,10^{-6}\}$ to stabilize score learning.

The model is optimized using Adam with learning rate $5\times10^{-5}$ and $(\beta_1,\beta_2)=(0.9,0.999)$. A multi-step scheduler reduces the learning rate by a factor of $0.1$ after 4000 steps. Training runs for 100 epochs with batch size 8 on NVIDIA GeForce RTX 3090 GPUs (24GB), typically consuming 3--6GB of memory per GPU. Under this setup, training completes in approximately 30 minutes on a multi-GPU workstation.

\subsection{Micrograph Curation Settings}

At inference time, the trained model is applied to full cryo-EM micrographs for denoising-based micrograph curation. We run five refinement iterations with a spatially varying noise map defined as $\sigma(x)=a+bx$, where $a=0.5$ and $b=0.01$ by default (slightly adjusted across datasets). Each iteration updates the current estimate using the predicted score field.

Inference is conducted on NVIDIA GeForce RTX 3090 GPUs. Due to the large micrograph resolution, processing full images may require up to $\sim$22GB of GPU memory. Processing an entire dataset typically takes several hours depending on the number and resolution of micrographs.

\section{Downstream Evaluation Details}
\subsection{CryoSPARC Configuration.}
We evaluated downstream particle picking and 3D reconstruction on three datasets: EMPIAR-10081, EMPIAR-10289, and EMPIAR-10291. Motion-corrected MRC files generated by CryoPPP were imported into CryoSPARC, followed by \texttt{Patch CTF Estimation} for each micrograph. Particle coordinates obtained from TOPAZ-based picking were then used in the \texttt{Extract from Micrographs} job with a box size of 256 pixels. The extracted particles were subsequently processed using \texttt{2D Classification} and \texttt{2D Select}, followed by ab-initio reconstruction and \texttt{Non-Uniform Refinement}. For EMPIAR-10289, we additionally performed \texttt{Homogeneous Refinement} before \texttt{Non-Uniform Refinement}.

During ab-initio reconstruction and refinement, we used symmetries C4, C8, and C8 for EMPIAR-10081, EMPIAR-10289, and EMPIAR-10291, respectively. All other CryoSPARC settings were kept at their default values. We used CryoSPARC v4.7.1~\cite{punjani2017cryosparc}.
The numbers of picked particles and selected particles used for reconstruction are reported in Tables~\ref{tab:picking_summary}.
All CryoSPARC jobs were run on NVIDIA Tesla V100 GPUs with 32\,GB memory per GPU and AMD EPYC 7742 CPUs.

\begin{table}[t]
\centering
\resizebox{\textwidth}{!}{
\begin{tabular}{c  c c c c c c c}
\toprule
EMPIAR ID & Original & Gaussian & DRACO & TOPAZ & DSM & TSM & $\#$ of True Protein Particles\\
\midrule
10081 & 53,247  & 53,260  & 46,876  & 46,386  & 52,438  & 49,674 & 39,352 \\
10289 & 67,190  & 67,213  & 75,960  & 75,377  & 69,997  & 80,506  & 61,517 \\
10291 & 109,725 & 116,239 & 148,909 & 147,220 & 122,154 & 130,041 & 99,808 \\
\bottomrule
\end{tabular}
}
\caption{Summary of the number of picked particles for the selected datasets.}
\label{tab:picking_summary}
\end{table}

\section{Additional Results}

\subsection{Particle Picking Evaluation}

\begin{table}[t]
\centering
\begin{tabular}{c|c|c|c}
\toprule
Method 
& EMPIAR-10291 
& EMPIAR-10289
& EMPIAR-10081 \\
\midrule
Original 
& 0.611 $\pm$ 0.103 
& 0.348 $\pm$ 0.067 
& 0.528 $\pm$ 0.125 \\
\midrule
Gaussian 
& 0.605 $\pm$ 0.103 
& 0.348 $\pm$ 0.067 
& 0.528 $\pm$ 0.125 \\
\midrule
DRACO    
& 0.537 $\pm$ 0.102 
& 0.377 $\pm$ 0.085 
& 0.369 $\pm$ 0.152 \\
Topaz    
& 0.534 $\pm$ 0.103 
& \textbf{0.378 $\pm$ 0.078} 
& 0.374 $\pm$ 0.153 \\
\midrule
DSM      
& \textbf{0.596 $\pm$ 0.104} 
& 0.355 $\pm$ 0.073 
& 0.546 $\pm$ 0.135 \\
TSM 
& 0.585 $\pm$ 0.105 
& 0.377 $\pm$ 0.073 
& \textbf{0.556 $\pm$ 0.138} \\
\bottomrule
\end{tabular}
\caption{Macro-averaged precision (mean $\pm$ std across micrographs) under distance-based evaluation.}
\label{tab:macro_precision}
\end{table}

\begin{table}[t]
\centering
\begin{tabular}{c|c|c|c}
\toprule
Method 
& EMPIAR-10291 
& EMPIAR-10289
& EMPIAR-10081 \\
\midrule
Original 
& 0.656 $\pm$ 0.085 
& 0.291 $\pm$ 0.050 
& 0.760 $\pm$ 0.102 \\
\midrule
Gaussian 
& 0.685 $\pm$ 0.076 
& 0.291 $\pm$ 0.050 
& 0.760 $\pm$ 0.102 \\
\midrule
DRACO    
& 0.727 $\pm$ 0.047 
& 0.349 $\pm$ 0.051 
& 0.448 $\pm$ 0.128 \\
Topaz    
& 0.715 $\pm$ 0.045 
& 0.350 $\pm$ 0.046 
& 0.451 $\pm$ 0.135 \\
\midrule
DSM      
& 0.703 $\pm$ 0.068 
& 0.306 $\pm$ 0.041 
& \textbf{0.770 $\pm$ 0.095} \\
TSM 
& \textbf{0.729 $\pm$ 0.058} 
& \textbf{0.362 $\pm$ 0.040} 
& 0.748 $\pm$ 0.100 \\
\bottomrule
\end{tabular}
\caption{Macro-averaged recall (mean $\pm$ std across micrographs) under distance-based evaluation.}
\label{tab:macro_recall}
\end{table}

Tables~\ref{tab:macro_precision} and \ref{tab:macro_recall} further show that DSM and TSM improve particle picking in different ways. DSM mainly leads to stronger recall gains across datasets, with the clearest improvement on EMPIAR-10081. This suggests that DSM is particularly helpful for recovering weak particle signals and making particles easier to detect.

TSM shows a different advantage. While keeping recall at a similarly high level, it achieves stronger precision on the more challenging datasets. In particular, TSM gives the best recall on EMPIAR-10289 and the best precision on EMPIAR-10081. This more balanced behavior is also reflected in its stronger overall F1 performance across datasets.

\subsection{3D Reconstruction Evaluation}

\begin{table}[h]
\centering
\resizebox{\textwidth}{!}{
\begin{tabular}{c c c c c c c c}
\toprule
EMPIAR ID & Original & Gaussian & DRACO & TOPAZ & DSM & TSM & GT \\
\midrule
10081 & \textbf{4.037} & 4.047 & 6.389 & 7.351 & 4.125 & 4.097 & 3.500 \\
10289 & 4.312 & 4.335 & 5.907 & 6.334 & 6.503 & \textbf{4.308} & 3.800 \\
10291 & 6.607 & 3.848 & 6.079 & 4.678 & 3.778 & \textbf{3.754} & 3.600 \\
\bottomrule
\end{tabular}
}
\caption{3D reconstruction resolution (\AA) for the selected datasets. Lower is better.}
\label{tab:3D_resolution_summary}
\end{table}

\begin{figure}[h]
    \centering
    \includegraphics[width=1\linewidth]{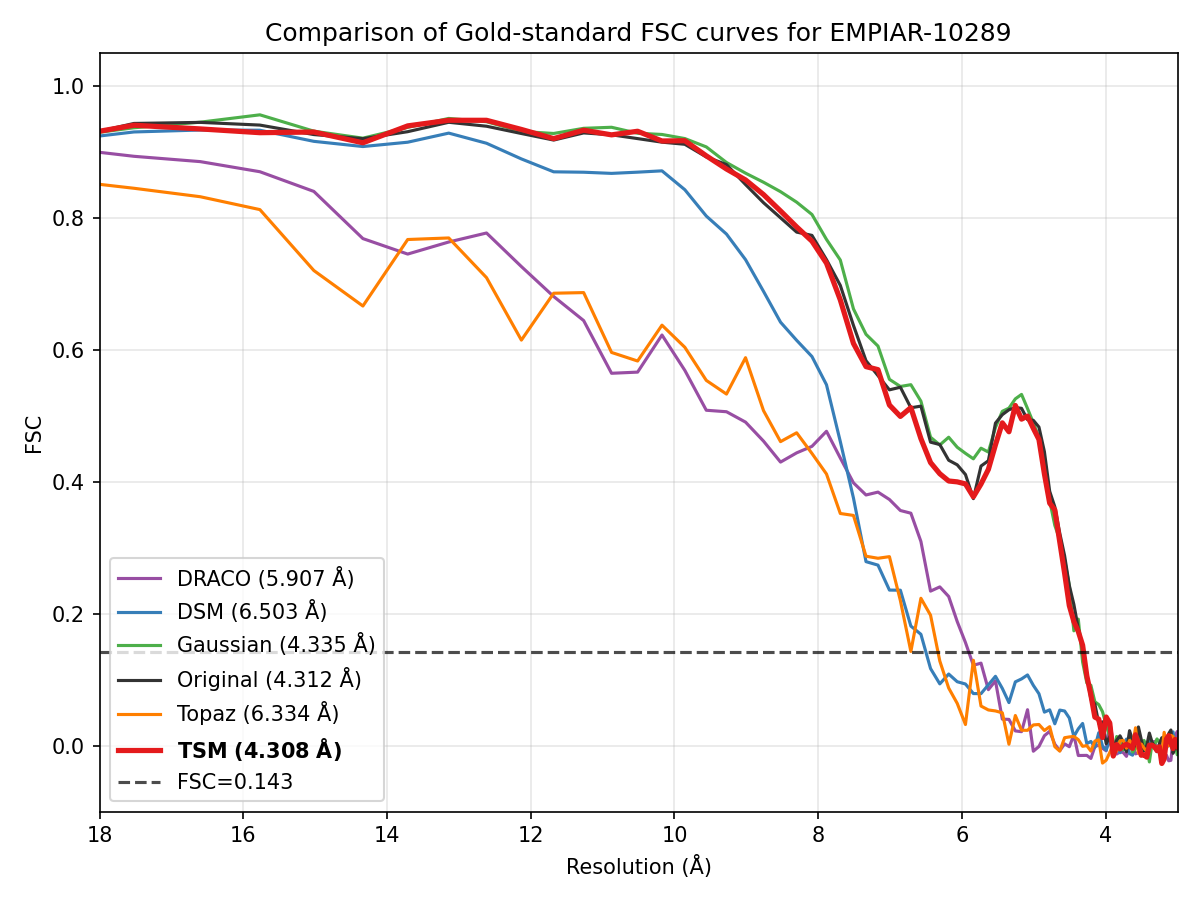}
    \caption{FSC curves for 3D reconstructions on EMPIAR-10289. Resolution is determined at FSC $=0.143$; a right-shifted curve indicates better resolution.}
    \label{fig:10289_fsc}
\end{figure}

Table~\ref{tab:3D_resolution_summary} shows that TSM achieves the best or near-best 3D reconstruction resolution across the three EMPIAR datasets. In particular, TSM gives the best resolution on EMPIAR-10289 and EMPIAR-10291, while remaining highly competitive on EMPIAR-10081. The differences between the strongest methods are within approximately $0.1\,\text{\AA}$, suggesting that their reconstruction quality is comparable. Overall, these results indicate that the benefits of TSM are preserved in downstream reconstruction and are not limited to particle picking alone.

This trend is further supported by the FSC curves in Fig.~\ref{fig:10289_fsc}. On EMPIAR-10289, the reconstruction based on TSM reaches the FSC $0.143$ threshold at the highest spatial frequency among the compared methods, corresponding to the best final resolution in Table~\ref{tab:3D_resolution_summary}. Overall, these results suggest that TSM improves particle quality in a way that benefits both downstream particle localization and 3D structural recovery.

\section{Additional Ablation Study Details}

This section provides additional qualitative comparisons that complement the ablation experiments presented in Sec.~\ref{ablation_sensitivity} of the main paper. 
In particular, we show full micrograph denoising results to illustrate how different design choices affect structural recovery in cryo-EM images.

\subsection{Effect of Target Weight $w_t$}

\begin{figure}[t]
    \centering
    \setlength{\tabcolsep}{2pt}
    \renewcommand{\arraystretch}{1.0}

    \begin{tabular}{c c}
        \includegraphics[width=0.49\textwidth]{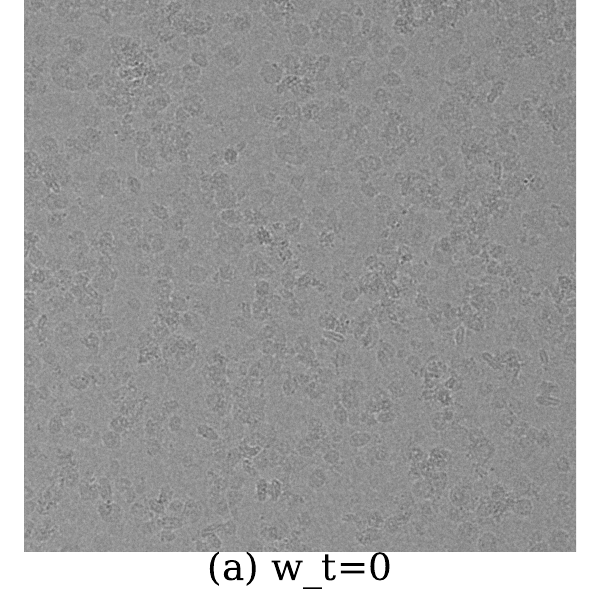} &
        \includegraphics[width=0.49\textwidth]{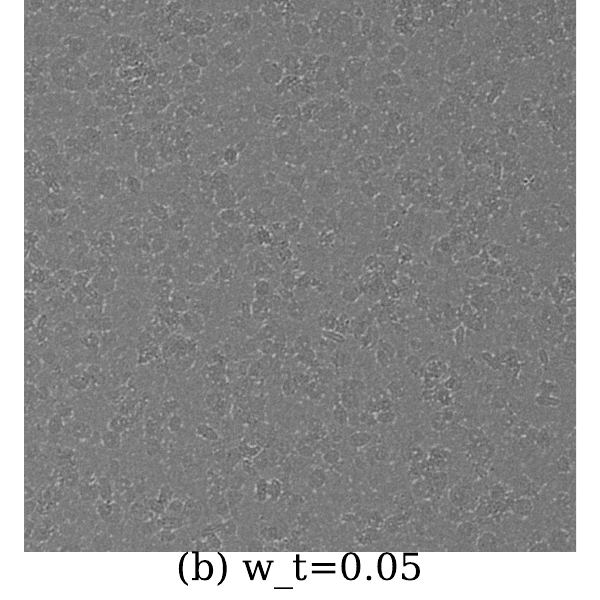}\\
        \includegraphics[width=0.49\textwidth]{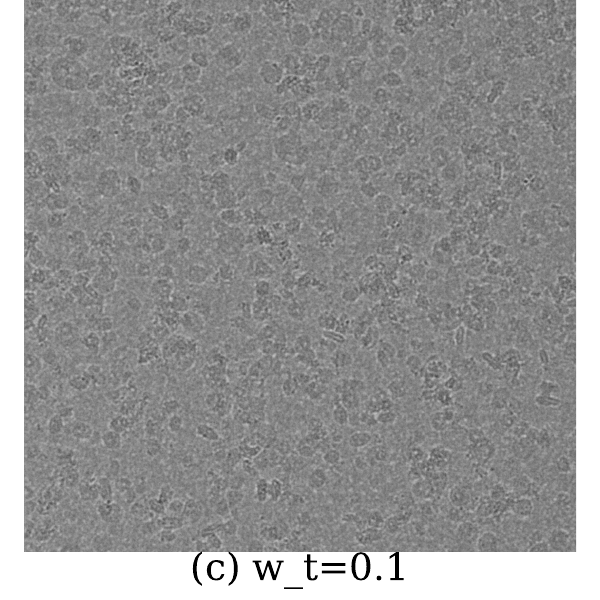} &
        \includegraphics[width=0.49\textwidth]{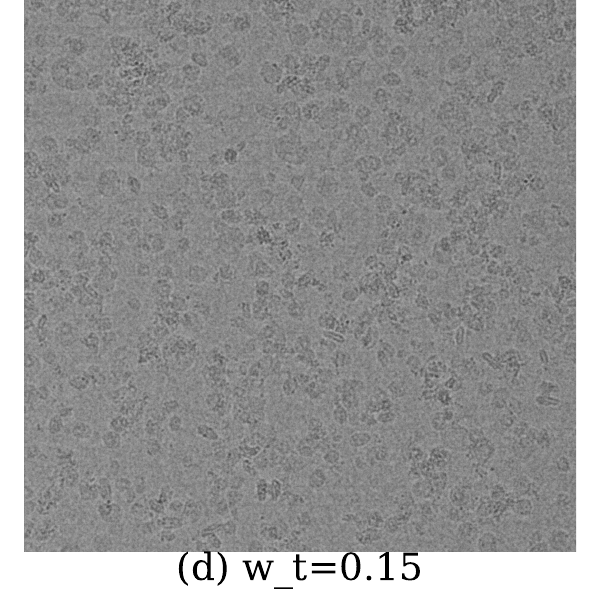} \\
    \end{tabular}

    \caption{
    Qualitative comparison of denoised micrographs under different target weights $w_t$ on EMPIAR-10289: $w_t=0$ (DSM only), $w_t=0.05$, $w_t=0.1$, and $w_t=0.15$. 
    Introducing target guidance noticeably improves particle boundary clarity and suppresses background noise. 
    Moderate values of $w_t$ provide the best balance between structural enhancement and preservation of fine details.
    }
    \label{fig:sup_wt_micrograph}
\end{figure}

\begin{figure}[t]
    \centering
    \setlength{\tabcolsep}{2pt}
    \renewcommand{\arraystretch}{1.0}

    \begin{tabular}{c c}
        \includegraphics[width=0.49\textwidth]{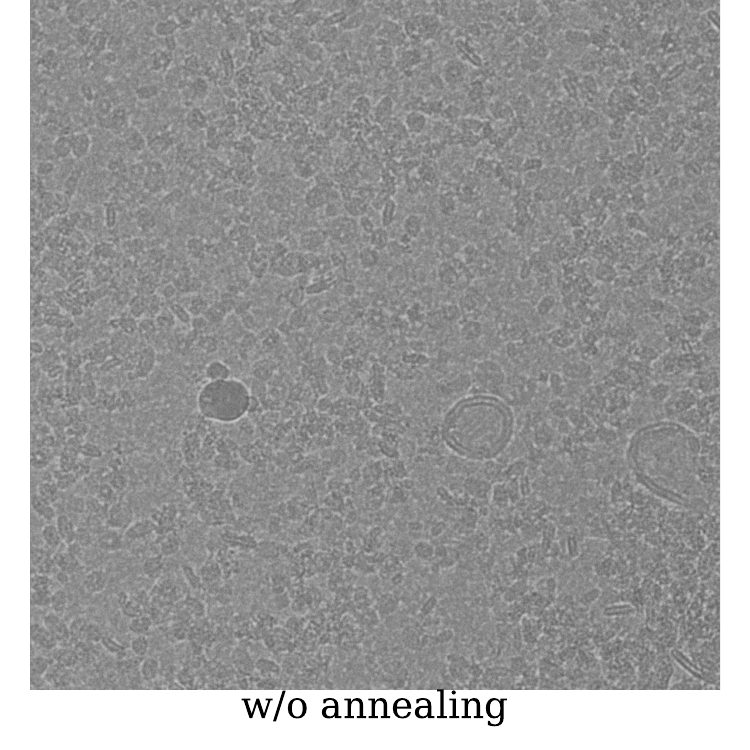} &
        \includegraphics[width=0.49\textwidth]{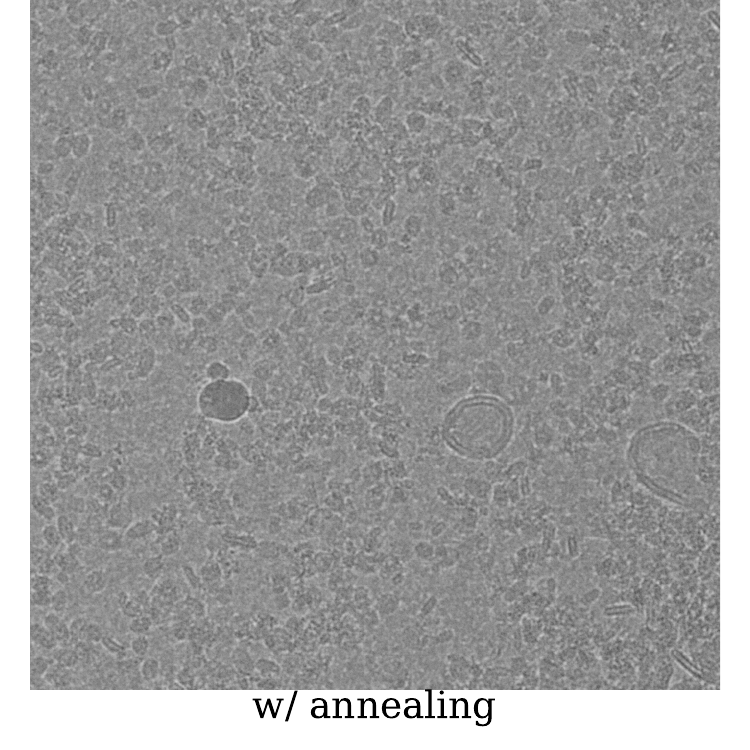}\\
    \end{tabular}

    \caption{
    Qualitative comparison of denoised micrographs with and without annealed global guidance on EMPIAR-10289. 
    Left: training without annealing. 
    Right: training with annealed learning schedule. 
    The annealed strategy produces more consistent particle structures and reduces residual background artifacts.
    }
    \label{fig:sup_anneal_micrograph}
\end{figure}

To further illustrate the impact of target guidance, Fig.~\ref{fig:sup_wt_micrograph} shows denoising results for the same micrograph under different values of the target weight $w_t$. 
When $w_t=0$, the model reduces noise but produces relatively blurred particle structures. 
Enabling TSM ($w_t>0$) leads to clearer particle boundaries and improved structural consistency across the micrograph. 
As discussed in the main paper, performance remains stable across a range of moderate values, while excessively large weights may introduce slight over-smoothing.

\subsection{Effect of Annealed Global Guidance}

Fig.~\ref{fig:sup_anneal_micrograph} presents a visual comparison of denoising results obtained with and without the annealed guidance. 
Without annealing, target guidance is applied from the beginning of training, when feature representations are still unstable, which can lead to inconsistent structural responses across the micrograph. 
In contrast, annealed learning delays the introduction of target guidance until the encoder has learned more stable representations. 
This strategy produces more uniform particle structures and cleaner background regions, consistent with the improved optimization stability discussed in the main text.